\documentclass[conference]{IEEEtran}
\usepackage[T1]{fontenc}
\usepackage[utf8]{inputenc}
\IEEEoverridecommandlockouts
\usepackage{amsmath,amssymb,amsfonts,mathtools,amsthm}
\usepackage{graphicx}
\usepackage{textcomp}
\usepackage{xcolor}
\usepackage{booktabs}
\usepackage{algorithm}
\usepackage{algpseudocode}
\usepackage{bm} 
\usepackage[caption=false,font=footnotesize]{subfig}
\usepackage{float}        % for [H] exact placement of single-column floats
\usepackage[section]{placeins} % keep floats within their section
\usepackage{dblfloatfix}  % better control over two-column floats (figure*)
\usepackage{cite}
\usepackage[colorlinks=true, linkcolor=black, citecolor=black, urlcolor=black]{hyperref}

\def\BibTeX{{\rm B\kern-.05em{\sc i\kern-.025em b}\kern-.08em
    T\kern-.1667em\lower.7ex\hbox{E}\kern-.125emX}}

\begin{document}

\IEEEpubid{\begin{minipage}[t]{\textwidth}\ \\[8pt]%
    \centering\footnotesize%
    This is an extended version of the paper published in the Proceedings of the 2025 IEEE International Conference on Machine Learning and Applications (ICMLA 2025).%
\end{minipage}}

\title{Enhancing Multi-Agent Collaboration with Attention-Based Actor-Critic Policies}

\author{
\IEEEauthorblockN{1\textsuperscript{st} Hugo Garrido-Lestache Belinchon}
\IEEEauthorblockA{\textit{Department of Computer Science and Software Engineering} \\
\textit{Milwaukee School of Engineering}\\
Milwaukee, WI USA \\
garrido-lestacheh@msoe.edu}
\and
\IEEEauthorblockN{2\textsuperscript{nd} Jeremy Kedziora}
\IEEEauthorblockA{\textit{Department of Computer Science and Software Engineering} \\
\textit{Milwaukee School of Engineering}\\
Milwaukee, WI USA \\
kedziora@msoe.edu}
}

\maketitle
\IEEEpubidadjcol
%%%%%%%%%%%%%%%%%%%%%%%%%%%%%%%%%%%%%%%%%%%%%%%%%%%%%%%%%%%%%%%

%@@@@@@@@@@@@@@@@@@@@@@@@@@@@@@@@@@@@@@@@@@@@@@@@@
\begin{abstract}

This paper introduces Team-Attention-Actor-Critic (TAAC), a reinforcement learning algorithm designed to enhance multi-agent collaboration in cooperative environments.
TAAC employs a Centralized Training/Centralized Execution scheme incorporating multi-headed attention mechanisms in both the actor and critic. 
This design facilitates dynamic, inter-agent communication, allowing agents to explicitly query teammates, thereby efficiently managing the exponential growth of joint-action spaces while ensuring a high degree of collaboration. 
%We further introduce a penalized loss function which promotes diverse yet complementary roles among agents.
We evaluate TAAC in a diverse range of cooperative environments and benchmark its performance against other multi-agent paradigms, including Proximal Policy Optimization (PPO) and Multi-Actor-Attention-Critic (MAAC).
We find that TAAC outperforms other algorithms as the level of coordination and collaboration required increases and that it performs comparably to the other algorithms in environments which require less coordination and collaboration.
\end{abstract}


%%%%%%%%%%%%%%%%%%%%%%%%%%%%%%%%%%%%%%%%%%%%%%%%%%%%%%%%%%
\section{Introduction}
\noindent When human organizations address a large cooperative problem or a task, they scale the number of individuals at the expense of having to organize and coordinate them.
Over thousands of generations, we have honed our ability to organize many individuals to collaborate at scale to reach a common goal, and woven it into our largest architectural and organizational achievements. As we integrate reinforcement learning into real-world applications, we need multi-agent systems that efficiently scale collaborative action to large numbers of agents to address broader needs and problems. In this paper, we study methods to manage cooperative multi-agent environments or systems. 
Our goal is to teach groups of AI agents strong collaborative skills that allow them to act as a team in a scalable fashion, thereby tackling complex problems and challenges efficiently.

The best methods for training a team to collaborate in a cooperative environment remain an open question.
The main paradigms that have been used to learn collaboration across agents in cooperative MARL are Decentralized Training/Decentralized Execution (DTDE),  Centralized Training/Decentralized Execution (CTDE), and Centralized Training/Centralized Execution (CTCE). In DTDE, agents operate independently without shared information both during training and inference while in CTDE, information is shared during training (usually to estimate values) but not when agents select actions or during inference. Finally, CTCE shares information during execution, which allows delegation of control to a single controller that determines the joint-action of all agents. DTDE and CTDE schemes present challenges due to the need to decide whether and how to share information across agents; these modeling choices are critical for team collaboration.\footnote{Decentralized execution schemes can also suffer from convergence problems due to the inherent non-stationarity that arises when multiple agents learn and adapt concurrently.} The CTCE single controller avoids these difficulties but poses another problem: the full joint-action space under a CTCE scheme is the product set of individual agent action sets and so its size grows exponentially in the number of agents. Effective team collaboration would seem to hinge on a robust search of the joint-action space for complementary actions across agents, and so a CTCE scheme imposes scalability constraints on collaborative activity.

%REWRITTEN CONTRIBUTION PARA (V2, Jeremy)
To address these challenges, we introduce and test a MARL algorithm that leverages shared information to foster inter-agent collaboration. Our proposed model, Team-Attention-Actor-Critic (TAAC), is a CTCE scheme that incorporates an attention mechanism directly into the actor to model an information-sharing ``conversation" between multiple agents which enables them to dynamically query teammates during decision-making. This allows our agents to manage the joint-action space in a scalable fashion. We evaluate TAAC in a diverse set of collaborative environments and benchmark its performance against PPO (as an example of DTDE) and MAAC (as an example of CTDE).
%We further introduce a loss that encourages agents to learn diverse yet complementary roles, enhancing collaboration. 

%%%%%%%%%%%%%%%%%%%%%%%%%%%%%%%%%%%%%%%%%%%%%%%%%%%%%%%%%%
\section{Related Work}
% justification for Centralised
\noindent In cooperative MARL environments, CTCE seems a natural choice to maximize collaborative potential, especially given the robustness of modern information sharing capabilities which would seem to allow for relatively costless message sending where agents either do not wish to or cannot restrict information sharing. A number of recent papers on MARL have proposed managing agent collaboration by leveraging modern deep learning architectures. Within the CTCE paradigm, \cite{Wen2022} and \cite{Liu2024} break down the full joint-action space by requiring agents to select actions sequentially and treating MARL as a sequence modeling problem by applying transformers.  \cite{Tavakoli2019} and \cite{info15050279} apply attention to deal with high-dimensional action spaces in a value-based context (to variants of deep $Q$-learning). By contrast, we focus on simultaneous action selection in a policy-based algorithmic context.\footnote{\cite{Wen2022} and \cite{Liu2024} would converge to a Markov Perfect Equilibrium for a stochastic game with within-period sequential actions by the agents; there are no guarantees that the equilibrium set for such a game would map well onto the equilibrium set for a game with true simultaneous choices.}  

Numerous CTDE schemes have explored different approaches to deal with information within multi-agent environments by modeling communication and information sharing between agents via a combination of heuristic rules (e.g. \cite{yang2020meanfieldmultiagentreinforcement}, \cite{sukhbaatar2016learningmultiagentcommunicationbackpropagation}), and explicitly modeling message selection (e.g. \cite{foerster2016learningcommunicatedeepmultiagent}), message timing (e.g. \cite{DOC} extending \cite{optioncritic}), or message recipients (e.g. \cite{liu2019multiagentgameabstractiongraph}, \cite{jiang2020graphconvolutionalreinforcementlearning}). Beyond these, some CTDE algorithms have applied attention to model communication between agents, particularly \cite{jiang2018learningattentionalcommunicationmultiagent} and \cite{das2020tarmactargetedmultiagentcommunication} which use attention mechanisms to learn when and with whom to communicate.  The former models communication groups dynamically via attention gating. The latter models targeted communication using attention mechanisms, enabling message exchange between specific agents.  Finally and perhaps most relevant for our work, MAAC \cite{MAAC} learns a centralized critic with a soft attention mechanism that dynamically selects agents to attend to at each time step during training.

These CTDE methods are particularly appropriate for teaching agents to manage environments collaboratively in which information sending is problematic, e.g. because agents are physically separated, because messages are costly to send, or because there are incentives to maintain private information. At the same time, if information sending is not the main constraint\footnote{Examples of such environments might be sports teams, simulated environments, digital games, or any environment in which a slight time lag to build a full picture of observations across agents is acceptable.}
then the focus on decentralized execution leads to more complicated methods than might be needed\footnote{Predefined or heuristic communication architectures restrict communication and hence restrain potential collaboration among agents.  A potential critique of \cite{yang2020meanfieldmultiagentreinforcement} and \cite{sukhbaatar2016learningmultiagentcommunicationbackpropagation} is that use of an aggregative operation across agents has the potential to eliminate differences among neighboring agents and result in loss of important information.}, likely underestimates the extent to which collaborative action is possible, and relegates it to no more than an implicit, emergent property during inference.

%In this paper, we build upon \cite{MAAC}, which explored the use of attention mechanisms in the critic in actor-critic architectures.  They additionally utilized a counterfactual multi-agent baseline from \cite{COMA} to generate a better estimate of the return. 

In this paper, we build upon \cite{MAAC} and \cite{COMA}. We extend their work by incorporating an attention mechanism in the actor which allows agents to receive information in a selective manner.
This allows us to let agents collaborate explicitly during inference.
It functions by allowing an agent to pose an open-ended question to its peers. 
For example, in a soccer game scenario, a player with the ball may query, `Who should I pass to?' Agents which are close to the goal may be identified as optimal recipients, and their responses inform the passing decision and ultimately improve collaboration. 
We have named this model Team-Attention-Actor-Critic (TAAC).

\section{Background and preliminaries}
\noindent In reinforcement learning, control problems with multiple agents can be modeled as a Partially observable Markov game \cite{PartiallyObservableStochasticGames}, which is defined by:

\begin{itemize}
\item A set of $n$ agents;
\item A set of states $S$ that describe the current common environmental conditions facing the agents;
\item A collection of actions $A(s) = \prod_{i}A_i(s)$ for $s\in S$ where $A_i(s)$ are the actions agent $i$ can take in state $s$ and:
\begin{align*}
\vec{a} = \langle a_1,a_2,\hdots,a_n\rangle\in A(s);
\end{align*}
\item Transition probabilities $p(s'\mid \vec{a} ,s)$ for transitioning from state $s$ to state $s'$; 
\item A function:
\begin{align*}
\vec{r}(s^{\prime},\vec{a},s) = \langle r_i(s',\vec{a},s)\rangle_{i=1,\hdots,n}
\end{align*} that supplies the immediate rewards for each agent associated with this transition.
\end{itemize}

% \noindent Throughout this paper, for a generic vector $\vec{z}$, we will write $\vec{z}_{\backslash i}$ to mean $z$ with the $i$th dimension excluded:
% \begin{align*}
% \vec{z}_{\backslash i} = \langle z_1,\ldots,z_{i-1},z_{i+1},\ldots,z_n\rangle
% \end{align*}
% and we will write $\vec{z}_i$ to mean the pair $(z_i, \vec{z}_{\backslash i})$.

In environments that take place over a finite number of discrete periods $T$, the sequence of periods in which the agents participate is referred to as an episode.  At each time step, each agent makes an observation, $o_{i,t} = O_i(s_t)$, of the global state, which encapsulates all available information for agent  $i$; the vector of such observations across agents is $\vec{o}_t = \langle o_{i,t}\rangle_{i=1,\hdots,n}$.  The goal of each agent is to learn a policy which describes the probability that a trained agent should take action $a_i$ given an observation $o_{i,t}$, to maximize its individual sequence of rewards throughout an episode: $\sum_{t = 0}^{T-1}\gamma^tr_i(s_{t+1},\vec{a}_t,s_t)$.  Here $\vec{a}_t$ and $s_t$ are the agents' actions and the common state at time $t$ and $\gamma\in[0,1]$ is the discount factor on future rewards.  In general, policies could differ across agents; we will focus on the case where the policy is shared.  Any differences in decision-making across agents are modeled using their observations of the state. Given this, we write these policies as $\pi(a_i|o_{i,t})$.

%@@@@@@@@@@@@@@@@@@@@@@@@@@@@@@@@@@@@@@@@@@@@@@@@@
\subsection{Policy Gradients}
\noindent Policy gradient methods model this policy directly, usually via a function approximator with parameters $\mathbf{u}$; we indicate the dependence of the policy on these parameters by writing $\pi_{\mathbf{u}}(\cdot)$.
%\footnote{By contrast, value-based methods (e.g. $Q$-learning, \cite{Q-learning}) estimate the state-conditional value of actions and then use those values to build a policy.} 
To learn $\mathbf{u}$ for $\pi_{\mathbf{u}}(\cdot)$, these methods use an estimate of the policy gradient (\cite{PolicyGradientTheorem}, \cite{williams-1988}, \cite{williams-1992}) to update the parameters in the direction of increase in the expected sum of discounted rewards. In the case of $n$ agents with shared policies, taking the Monte Carlo approach of using the full return, the policy gradient may be written as: 
\begin{equation}
\nabla_{\mathbf{u}} J(\mathbf{u}) = \mathbb{E}_{\tau} \Biggl[ \sum_{i=1}^{n} \sum_{t=0}^{T-1} \nabla_{\mathbf{u}} \ln \pi_{\mathbf{u}} \bigl(a_{i,t} \mid o_{i,t} \bigr) \, [G_{i,t}(\tau) - b(o_{i,t})] \Biggr],
\end{equation}
where $\tau$ is a trajectory of states and actions whose distribution is dependent on $\pi_{\mathbf{u}}(\cdot)$ and on unknown environmental dynamics:
\begin{align*}
% \tau = s_0,\vec{a}_0,s_1,&\vec{r}(s_1,\vec{a}_0,s_0),\vec{a_1}\\
% &\ldots,s_{T-1},\vec{a}_{T-1},s_T,\vec{r}(s_T,\vec{a}_{T-1},s_{T-1}). %based on our previous definition of $\vec{z}_i$ each one means $(z_i, \vec{z}_{\backslash i})$, I don't think the maths break but not sure. please check it if you have time
\tau = s_0,\vec{a}_0,s_1,\vec{a}_1,\ldots,s_{T-1},\vec{a}_{T-1},s_T
\end{align*}
and $G_{i,t}(\tau)$ is the sum of discounted rewards derived from trajectory $\tau$ from a given timestep $t$ onwards (the causal return) for a given agent $i$:
\begin{equation}
G_{i,t}(\tau) = \sum_{k=t}^{T-1} \gamma^{\,k - t} \, r_i\bigl(s_{k+1}, \, a_k, \, s_k\bigr). 
\end{equation}
Finally, $b(\cdot)$ is a baseline subtracted from the return to avoid high variance in the estimates of $\nabla_{\mathbf{u}}J(\mathbf{u})$ across episodes driven by use of $G_{i,t}(\tau)$.\footnote{An arbitrary function of the state can be introduced into the policy gradient equation without changing the expectation -- see \cite{PolicyGradientTheorem} for details.  If $b(\cdot)$ is allowed to depend on agent actions then this will bias the estimate of the gradient.}

Given sampled episodes, this estimate of the policy gradient is used to take steps in the direction of $\nabla_{\mathbf{u}} J(\mathbf{u})$ via gradient ascent to reach an optimal policy.% which is shared by all agents

%@@@@@@@@@@@@@@@@@@@@@@@@@@@@@@@@@@@@@@@@@@@@@@@@@
\subsection{Actor-Critic Algorithms}
\noindent A common choice for the baseline function $b(\cdot)$ is an estimate of the long-term value, for example, the state-value function, $V(o_{i,t})$, which predicts the return to an agent given an observation $o_{i,t}$ (see \cite{suttonbarto2018}).  If we model the state value function as a differentiable function of parameters $\mathbf{w}$ and write $V_{\mathbf{w}}(\cdot)$ then this yields the following baseline:
%\begin{align*}
%\nabla_{\mathbf{u}} J(\mathbf{u})= \mathbb{E}_{\tau \sim \pi_{\mathbf{u}}} \Biggl[ \sum_{i=1}^{n} \sum_{t=0}^{T-1} %\nabla_{\mathbf{u}}& \ln \pi_{\mathbf{u}} \bigl(a_{i,t} \mid o_{i,t} \bigr)\\
%&\times\left[G_{i,t}(\tau) - V_{\mathbf{w}}(o_{i,t})\right] \Biggr].
%\end{align*}
\begin{align*}
b(o_{i,t}) &= V_{\mathbf{w}}(o_{i,t}) = \sum_{a\in A_i}\pi_{\mathbf{u}}(a|o_{i,t})Q_{\mathbf{w}}(o_{i,t},a)
\end{align*}
where the second equality follows from the usual definitions of the state-value function $V_{\mathbf{w}}(\cdot)$ and the state-action value function $Q_{\mathbf{w}}(\cdot)$.\footnote{We use the Counterfactual Multi-agent baseline described in \cite{COMA} in the actors' policy gradients.}% We will use this form of the baseline below to model multi-agent structure.%Note that here the state-action value function is averaged over actions according to their realization under the policy, and so this choice of baseline does not depend upon actions.  

The policy is referred to as the actor and the long-term value estimate as the critic, and the resulting model is called an actor-critic algorithm.  The actor's parameters are learned through the policy gradient estimates and the critic's parameters are learned through value-based methods (e.g. gradient descent on the difference between the critic's predicted state value and a target value, which can be either the observed $G_{i,t}(\tau)$ or a bootstrapped value, see \cite{mnih2013}).

%%%%%%%%%%%%%%%%%%%%%%%%%%%%%%%%%%%%%%%%%%%%%%%%%%%%%%%%%%%%%%%%%%%%%%%%%%
\section{Team-Attention-Actor-Critic (TAAC)}
\subsection{Attention Mechanism}
%The authors of \cite{MAAC} introduced the use of an attention mechanism in the critic, the purpose of which is to aggregate information from multiple agents to more accurately evaluate an agent's state-action value. The intuition behind this use of the attention mechanism is to enable the critic to ``query" the agents for their internal representations, thereby enhancing the information available to optimize the critic and allowing for more accurate prediction of the return. 

\noindent We extend \cite{MAAC} by introducing a separate attention mechanism in the actor. 
The intuition behind doing this is to allow agents to use the internal representations of other agents directly when making choices, with the idea that enabling this is analogous to allowing agents to ``put themselves into the shoes" of their colleagues and thereby giving them insight into what other agents will choose. If they build an understanding of what their colleagues will choose, then they may be able to create collaboration by selecting actions that dovetail well.

In general, attention works as follows (see \cite{attentionneed}).  An initial input $z$ is suitably embedded as a sequence of vectors $M(z) = \{m_1(z), \hdots, m_n(z)\}$ and multiplied with learned matrices $W_Q$, $W_K$, and $W_V$. Attention is then computed as:
\begin{equation}
A(M(z)|W) = \text{Softmax}\left(\frac{M(z)W_Q(M(z)W_K)^T}{\sqrt{d_K}}\right)M(z)W_V
\end{equation}
where $W = (W_Q, W_K, W_V)$ and $d_K$ is the dimensionality of the $M(z)W_K$ matrix.\footnote{The subscripts refer to the usual presentation of $W_Q$, $W_K$, and $W_V$ as the query, key, and value matrices, respectively.  Note that in all other sections of the paper $Q(\cdot)$ will refer to the state-action value function.}\footnote{Division by $d_K$ is done to scale the values for numerical stability.}

In a natural language application, the input might be a sequence of embedded tokens (e.g. words). In our application, these embedded input tokens will represent individual agent observation and/or action information.  We apply multiple attention heads in parallel, each with its own set of weight matrices.  The updated output sequence of tokens from these heads are concatenated by agent to create an updated set of agent-specific embedding information. Specifically, given $h$ attention heads, for agent $i$ we gather updated agent-specific embedding information by concatenating:
\begin{align}
E_i(z|W) = \{A_{i,j}(M(z)|W_j)\}_{j=1}^{h}\label{eqn:concat}
\end{align}
where $A_{i,j}(\cdot|W_j)$ refers to the values in the $i$th row of the $j$th head's output.%and then passed through a linear transformation to achieve an updated set of agent-specific embedding information.

These tokens are individually forwarded through the rest of the architecture to generate either a distribution over actions (in the actor) or a state-action value estimate (in the critic); we describe both below.

%@@@@@@@@@@@@@@@@@@@@@@@@@@@@@@@@@@@@@@@@@@@@@@@@@
\subsection{Actor Attention Architecture}
\noindent We model the policy with multiple output layers that share parameters but take different inputs, one for each agent.  To incorporate attention into the policy, we condition it on the observations of all agents and write it as $\pi_\mathbf{u}(\vec{o})\in\prod_i\Delta(A_i)$ so that $\pi_{\mathbf{u}}(\vec{o})=\langle\pi^{(1)}_{\mathbf{u}}(\vec{o}),\hdots,\pi^{(n)}_{\mathbf{u}}(\vec{o})\rangle$.  Thus, $\pi_{\mathbf{u}}(\cdot)$ generates $n$ distributions, one over each action set $A_i$, and $\pi^{(i)}_{\mathbf{u}}(a_i|\vec{o})$ returns the probability that agent $i$ takes action $a_i\in A_i$.\footnote{We focus on the case where $A$ is discrete here.} The forward pass of the parameterized policy network is as follows.
\begin{enumerate}
\item For $i = 1,\hdots,n$ the observation $o_i$ is passed through an appropriate embedding layer, say $m^{\pi}_1(\cdot|u_{m^{\pi}_1})$, where $u_{m^{\pi}_1}$ are associated parameters, to generate initial embeddings which are formed into the matrix:
\begin{align*}
    M_{\pi}(\vec{o}|u_{m^{\pi}_1}) = \left(\begin{array}{c}
    m^{\pi}_1(o_1|u_{m^{\pi}_1})\\
    \vdots\\
    m^{\pi}_1(o_n|u_{m^{\pi}_1})
    \end{array}
    \right).
\end{align*}
\item These embeddings are passed into a layer of $h_{\pi}$ attention heads which yields: 
\begin{align*}
%A_j(M_a(\vec{o}|u_{m_1})|\mathbf{u}).
A_j(M_{\pi}(\vec{o}|u_{m^{\pi}_1})|u_{A_j}).
\end{align*}
where $u_{A_j} = (u_{A_j,K},u_{A_j,Q},u_{A_j,V})$ are the parameters for the $j$th actor attention head. 

\item For every agent $i=1,\ldots,n$ we gather updated agent-specific embedding information by concatenating as described in equation \ref{eqn:concat}, which we write in terms of the input observation vector as $E_i(\vec{o}|u_{m_1^\pi},u_A)$ where $u_A = (u_{A_1},\hdots,u_{A_{h_{\pi}}})$.

\item Each updated agent-specific embedding is passed through a set of MLP layers, say $m^{\pi}_2(\cdot|u_{m^{\pi}_2})$ where $u_{m_2^{\pi}}$ are associated parameters, and a final softmax operation is applied to generate $n$ distributions over actions, one for each agent.
\end{enumerate}

With all these steps in mind, we may write the output of the policy for the $i$th agent as:
\begin{align*}
\pi^{(i)}_{\mathbf{u}}(\cdot|\vec{o})= \mbox{Softmax}\left(m^a_2\left(m^a_1(o|u_{m_1^{\pi}}),
\left.E_i(\vec{o}|u_{m_1^{\pi}},u_A)\right|u_{m_2^{\pi}}\right)\right)
\end{align*}
where $\mathbf{u} = (u_{m_1^{\pi}},u_{A},u_{m_2^{\pi}})$.

\begin{figure}[!h]
    \centering
    \includegraphics[width=0.8\linewidth]{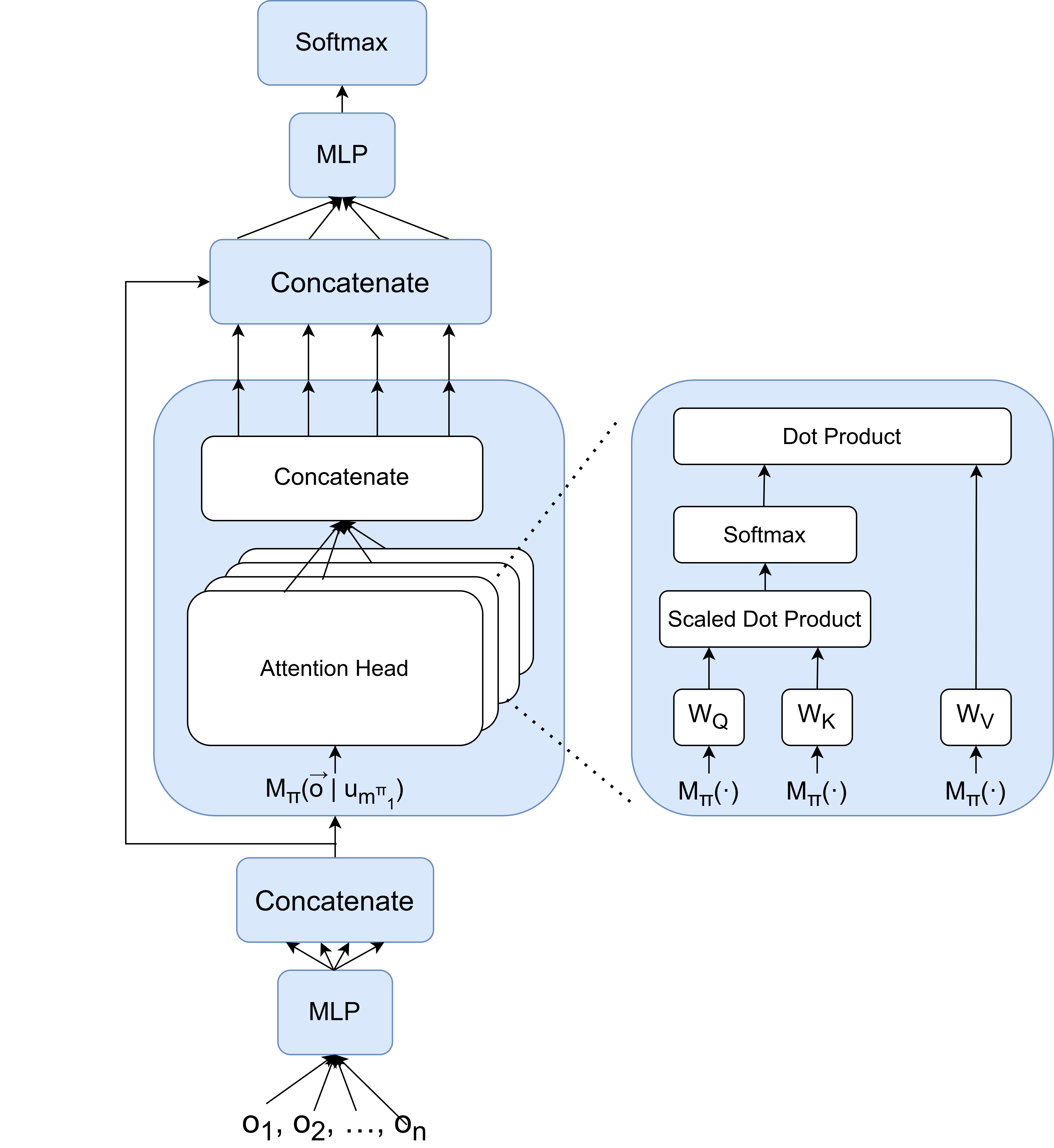}
    \caption{Architecture of Actor.  Note that the output is a set of $n$ distributions over actions, one for each agent.}
    \label{fig:actor}
\end{figure}

%@@@@@@@@@@@@@@@@@@@@@@@@@@@@@@@@@@@@@@@@@@@@@@@@@
\subsection{Critic Attention Architecture}
\noindent In the critic, we incorporate both the observations and actions of other agents. Consequently, the critic's state-action values are written as $Q_{\mathbf{w}}(\vec{o}, \vec{a})\in \mathbb{R}^n$ so that $Q_{\mathbf{w}}(\vec{o}, \vec{a}) = \langle Q^{(1)}_{\mathbf{w}}(\vec{o},\vec{a}),\hdots,Q^{(n)}_{\mathbf{w}}(\vec{o},\vec{a})\rangle$ and $Q^{(i)}_{\mathbf{w}}(\vec{o},\vec{a})$ provides that state-action value estimate for agent $i$ taking action $a_i\in A_i$.
%\begin{align*}
%%Q_{\mathbf{w}}\bigl((o_i, \vec{o}_{\backslash i}), (a_i, \vec{a}_{\backslash i})\bigr).
%Q_{\mathbf{w}}(\vec{o}_i, \vec{a}_i).
%\end{align*}
The forward pass of the parameterized critic network is analogous to that used for the policy; the initial embedding depends on parameters $w_{m_1^Q}$, there are $h_Q$ attention heads with $w_{A_j} = (w_{A_j,K},w_{A_j,Q},w_{A_j,V})$ the parameters for the $j$th critic attention head, the post-attention output is concatenated by agent as $E_i(\vec{o},\vec{a}|w_{m_1^Q},w_A)$ where $w_A = (w_{A_1},\hdots,w_{A_{h_Q}})$, and the output MLP layer uses parameters $w_{m_2^Q}$ to generate $n$ state-action value estimates, one for each agent. The output of the critic for the $i$th agent is:
\begin{align*}
Q^{(i)}_{\mathbf{w}}&(\vec{o},\vec{a})\\
&= m^Q_2\left(\left.m_1^Q\left(o_i,a_i|w_{m_1^Q}\right),E_i\left(\vec{o},\vec{a}_i|w_{m_1^Q},w_A\right)\right|w_{m_2^Q}\right)
\end{align*}
where $\mathbf{w} = (w_{m_1^{Q}},w_{A},w_{m_2^{Q}})$.

\begin{figure}[!h]
    \centering
    \includegraphics[width=0.8\linewidth]{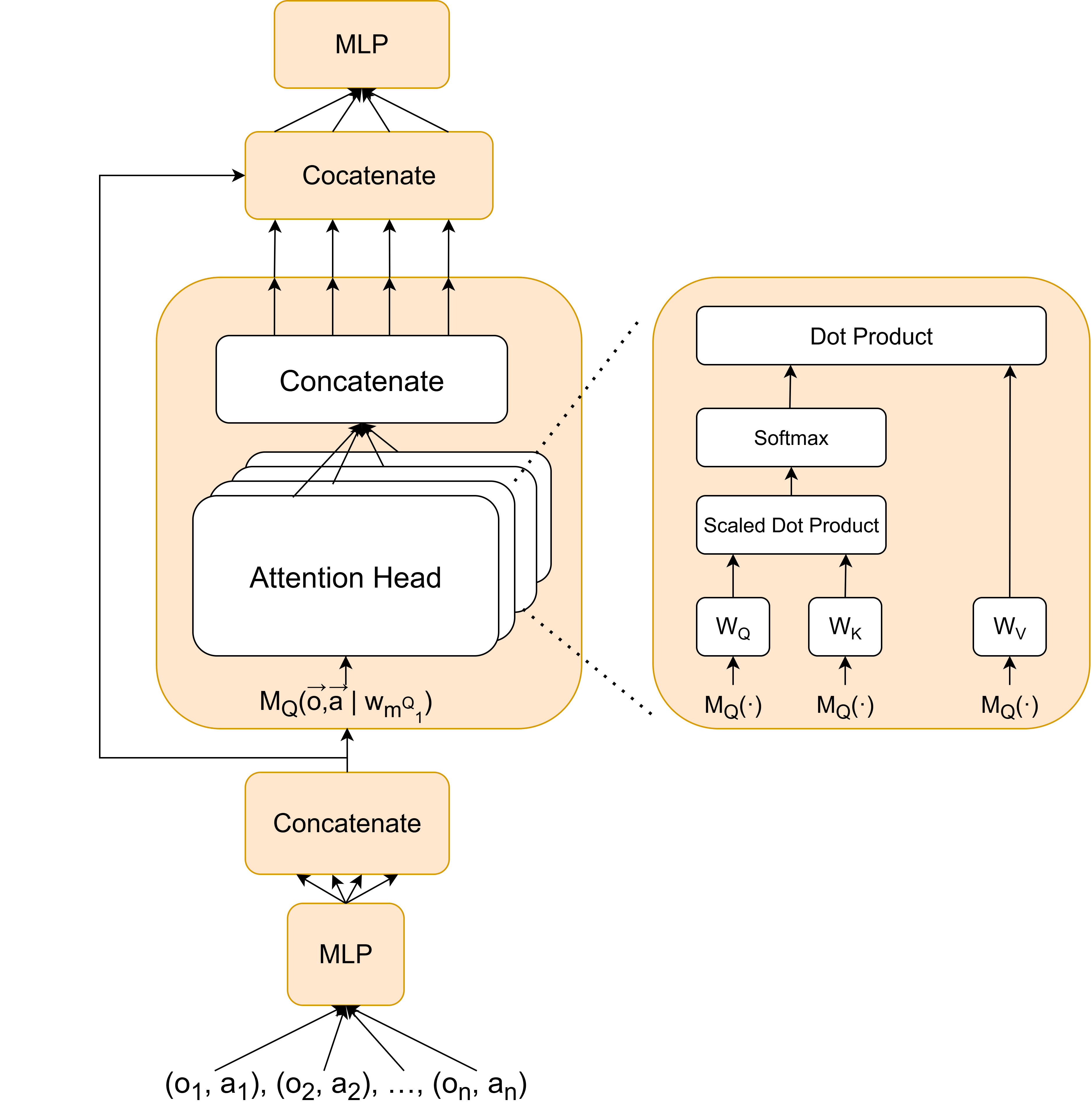}
    \caption{Architecture of Critic.  Note that the output is a set of $n$ state-action values, one for each agent.}
    \label{fig:critic}
\end{figure}

The actor and critic architectures are depicted in Figure~\ref{fig:actor} and Figure~\ref{fig:critic}, respectively.

%%%%%%%%%%%%%%%%%%%%%%%%%%%%%%%%%%%%%%%%%%%%%%%%%%%%%%%%%%%%%%%%
\section{Experiments}

% moved environments figure to appear with its descriptive text
\begin{figure*}[!t]
  \centering
  \subfloat[Soccer]{
    \includegraphics[width=0.40\textwidth]{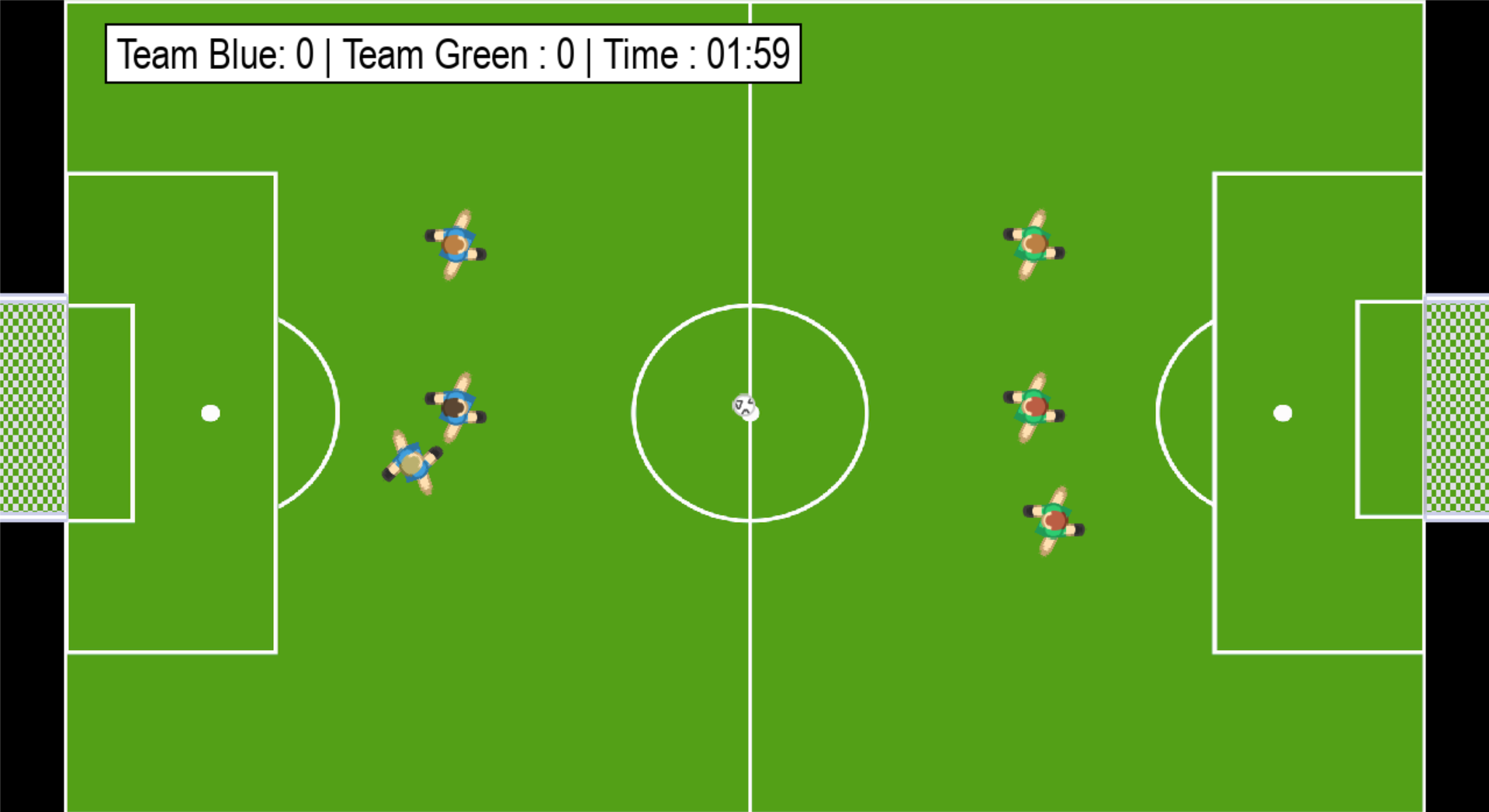}
    \label{fig:soccer}
  }\hfill
  \subfloat[BoxJump ~\cite{boxjump}]{
    \includegraphics[width=0.27\textwidth]{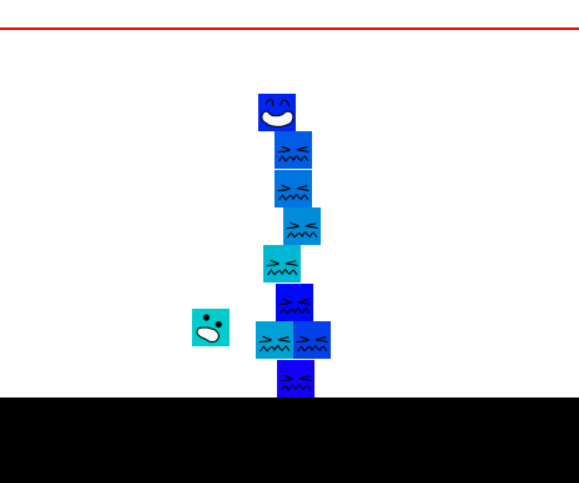}
    \label{fig:boxjump}
  }\hfill
  \subfloat[LBF~\cite{LBF1,LBF2}]{
    \includegraphics[width=0.23\textwidth]{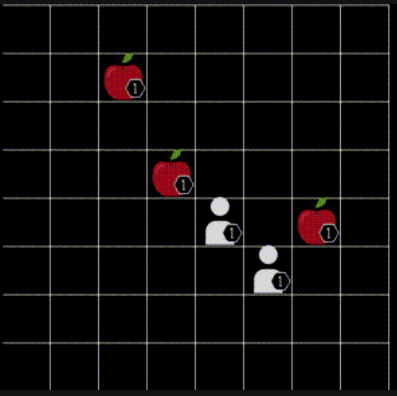}
    \label{fig:lbf}
  }
  \caption{Evaluation environments.} 
  \label{fig:envs}
\end{figure*}

\noindent We compare multiple algorithms against our novel architecture (TAAC) across the following environments: a novel simulated soccer environment, a cooperative tower-building environment (BoxJump) \cite{boxjump}, and Level-Based Foraging (LBF) \cite{LBF1,LBF2}. The soccer environment assesses performance in settings requiring high-level strategy and fine-grained action control; 
the BoxJump environment assesses performance with high numbers of agents; and the LBF environment evaluates performance on a simple, popular benchmark for cooperative MARL. In the soccer and LBF environments, agents observe the full state, thereby attributing performance improvements solely to coordination rather than to recovering missing state information. By contrast, observations in BoxJump are partial, allowing us to measure agents' capability to share local information at scale, testing the algorithm's ability to communicate/coordinate effectively and to gather only crucial information from other agents. Overall, these environments present a diverse set of cooperative multi-agent problems to assess the algorithms' performance. The soccer environment is fully described in \hyperref[Appendix:A]{Appendix A}, the BoxJump environment is fully described in \hyperref[Appendix:B]{Appendix B}, and the LBF environment is fully described in \cite{LBF1}. Figure~\ref{fig:envs} depicts the environments.

Our evaluation compares the performance of our TAAC algorithm against two benchmark algorithms: PPO \cite{PPO} and MAAC \cite{MAAC}. PPO was chosen because of its DTDE approach and its widespread use and proven robustness across diverse environments, while MAAC was chosen due to its CTDE approach and because it closely aligns with our implementation. Additionally, in the soccer environment, we included a random model to provide a grounded point of reference.

%@@@@@@@@@@@@@@@@@@@@@@@@@@@@@@@@@@@@@@@@@@@@@@@@@
\subsection{Training Procedures}
\noindent Below, we show describe our training methods to ensure that each model is trained to the best potential policy in the BoxJump and soccer environments. Due to LBF's simplicity, we did not implement any unique strategy for training.

\vspace{0.8em}
\noindent
\textit{A.1. BoxJump Training Procedure}
\vspace{0.2em}

\noindent To gradually guide the models toward the best policy, we employ curriculum learning \cite{curriculumlearningsurvey}. Given that the algorithm is not constrained by the number of agents, we can use curriculum learning to gradually increase the number of agents. To do this, at each episode, we randomly sample a number of agents; this becomes the number of agents used for that episode. Each curriculum stage uses a different agent-count range, as follows:
\begin{enumerate}
    \item \textbf{Stage 1:} Randomly select from [2, 3, 4, 5] agents for 30{,}000 episodes.
    \item \textbf{Stage 2:} Randomly select from [5, 6, 7, 8] agents for 20{,}000 episodes.
    \item \textbf{Stage 3:} Randomly select from [9, 10, 11, 12] agents for 2{,}000 episodes.
\end{enumerate}
All algorithms are trained for the same number of episodes in each stage; however, when moving to a new stage, we promote only the highest-reward policy. This procedure enables agents to first understand the reward structure and dynamics of the environment and then develop a collaborative policy.

\vspace{0.8em}
\noindent
\textit{A.2. Soccer Training Procedure}
\vspace{0.2em}

\noindent We also use curriculum learning \cite{curriculumlearningsurvey} to progressively build up skills in the soccer environment. Training is divided into the following stages:
\begin{enumerate}
    \item \textbf{Stage 1:} Score goals against an opposing team consisting of random agents, with random ball and player spawns.
    \item \textbf{Stage 2:} Engage in league play, where a team competes against a randomly selected prior version of itself, while maintaining random spawn positions for ball and players.
    \item \textbf{Stage 3:} Continue league play with fixed starting positions for each team.
\end{enumerate}
This staged learning approach allows agents to first master movement and scoring before tackling competitive play. 

League play ensures that the learning agent plays against a progressively more diverse and competent set of opponents. Our implementation of league play is a simplified version of \cite{vinyals2019}. For each episode we randomly sample opponents for the learning agent from a pool containing frozen past copies of agents trained using each of the possible algorithms, including an agent choosing actions randomly. During training, every 750 episodes, we add freeze a copy of the learning agent and add it to the pool.

%%%%%%%%%%%%%%%%%%%%%%%%%%%%%%%%%%%%%%%%%%%%%%%%%%%%%%%%%%%%%%%%
\section{Results}

%@@@@@@@@@@@@@@@@@@@@@@@@@@@@@@@@@@@@@@@@@@@@@@@@@
\subsection{BoxJump - Results}
\begin{figure}[!h]
  \centering
  \includegraphics[width=0.97\linewidth]{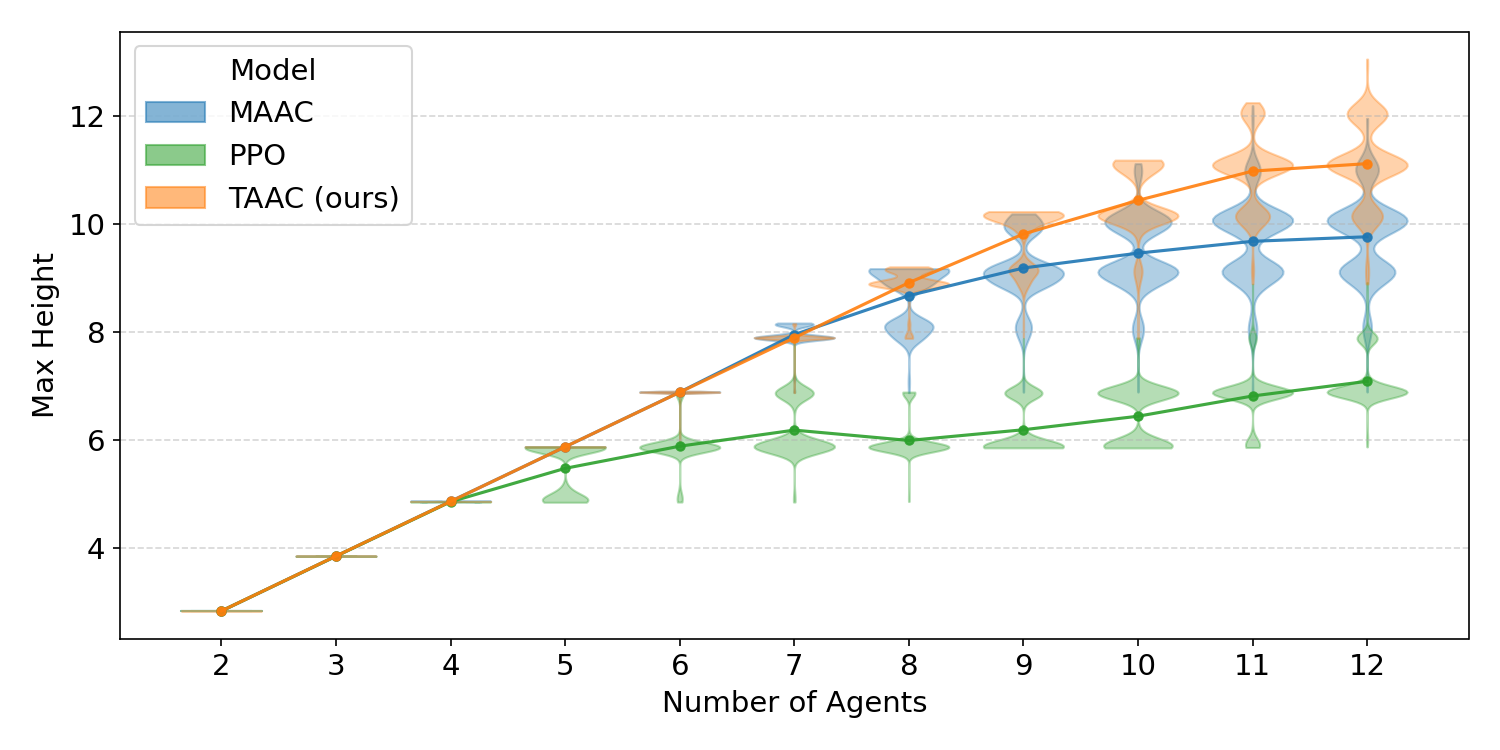}
  \caption{BoxJump: This figure showcases the maximum height achieved for each agent count and algorithm during inference. TAAC (ours) achieves more height as the number of agents increases, in expectation.
  A Kolmogorov-Smirnov Test and a Bootstrap test suggest that the distributions for MAAC and TAAC are significantly different at 8 agents and above.}
  \label{fig:boxjump_height}
\end{figure}

\noindent To evaluate BoxJump, we tested each algorithm at each agent count for 1000 episodes and gathered metrics.  For each agent count, we only selected the checkpoint of the algorithm which performed best for the agent count. Figure \ref{fig:boxjump_height} showcases the performance for each algorithm in the BoxJump environment. If TAAC is a superior approach for learning collaboration at scale then we would expect to see its relative performance increase as the number of agents increases.  Figure \ref{fig:boxjump_height} demonstrates this; TAAC-trained agents are likely to achieve a higher max height than alternative approaches at larger numbers of agents in expectation. The TAAC/alternative height distributions are significantly different at 8 agents and above. We provide more results in this environment in \hyperref[Appendix:C.1]{Appendix C.1}.

%\clearpage

%@@@@@@@@@@@@@@@@@@@@@@@@@@@@@@@@@@@@@@@@@@@@@@@@@
\subsection{LBF - Results}
\begin{figure}[!h]
  \centering
  \includegraphics[width=0.97\linewidth]{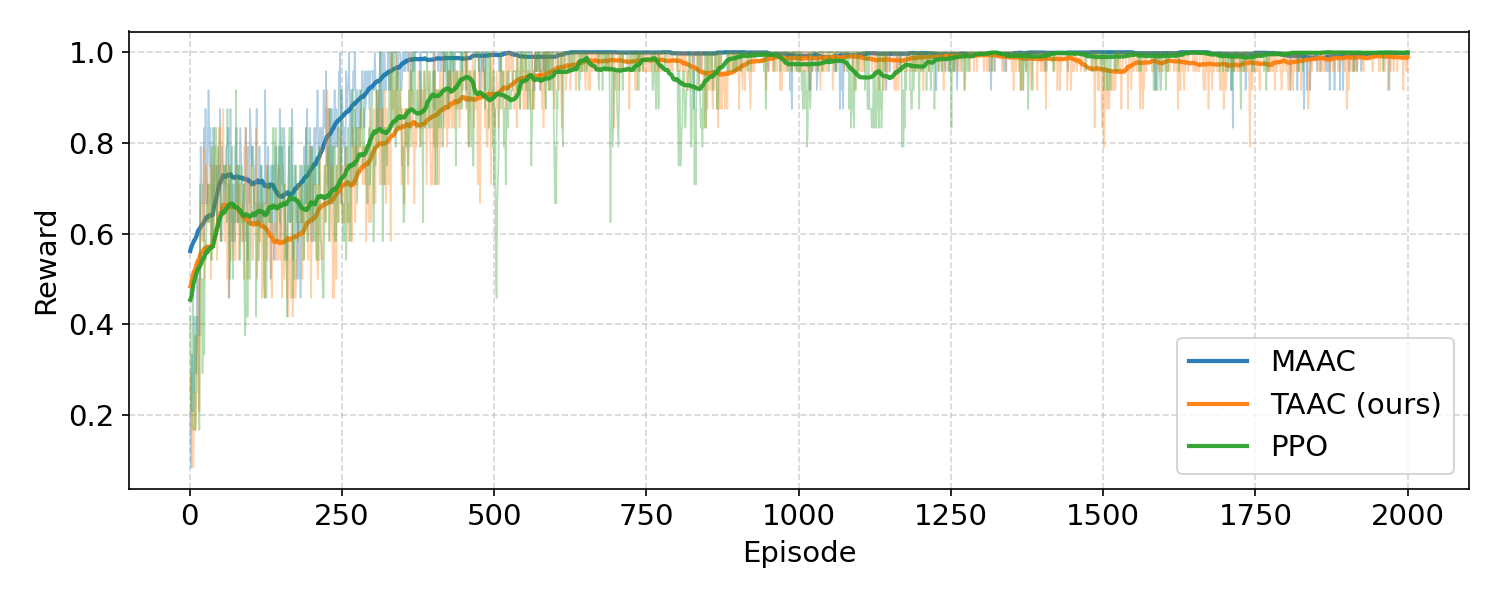}
  \includegraphics[width=0.97\linewidth]{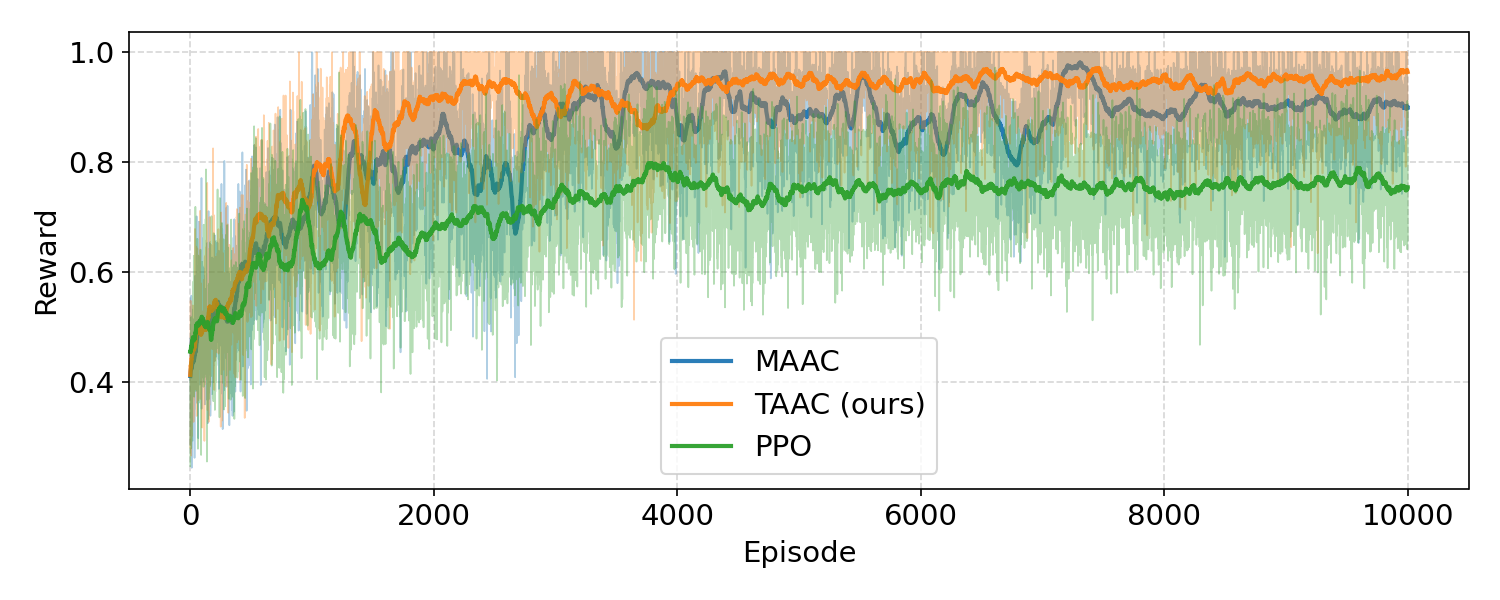}
  \caption{Top: LBF - Task: 2p-4f-5x5-c. Two players gather four food in a 5x5 grid world; each gathering requires all players. Bottom: LBF - Task: 3p-6f-6x6. Three players gather six food in a 6x6 grid world. Reward is normalized by food levels in both experiments.}
  \label{fig:lbf_results}
\end{figure}

\noindent In the LBF environment, we analyzed the performance and sample efficiency of TAAC vs alternatives across a range of different task complexity levels. Each task varies by the number of players (p), the number of food items to be collected (f), and the size of the board. They also vary by whether all agents are required to pick up each food, thus enforcing the need for more collaboration (c). If TAAC is a superior approach for learning collaboration then we would expect to see its relative performance increase as the collaboration becomes more necessary for task completion and as the difficulty of successfully doing it increases. Figure \ref{fig:lbf_results} measures the performance for each algorithm on two tasks in the LBF environment. The results suggest that: 1) TAAC has comparable sample efficiency and performance to the other algorithms in a simple task; and 2) TAAC has better performance with comparable sample efficiency in a more complex task which requires more collaboration. We provide results across all tasks in this environment in \hyperref[Appendix:C.2]{Appendix C.2}.

%@@@@@@@@@@@@@@@@@@@@@@@@@@@@@@@@@@@@@@@@@@@@@@@@@
\subsection{Soccer - Results}
\noindent Soccer is a more complicated environment which requires a high level of collaboration and strategy to succeed, mainly because it is partially adversarial. To gauge TAAC's performance in this environment, we trained it using the league play described above and then competed it against the alternative algorithms by randomly sampling opponents trained using those alternatives during evaluation episodes. We did this for 100,000 games and gathered a range of metrics to get an understanding of the performance and collaboration for each algorithm. 

\begin{figure}[!h]
  \centering
  \includegraphics[width=0.97\linewidth]{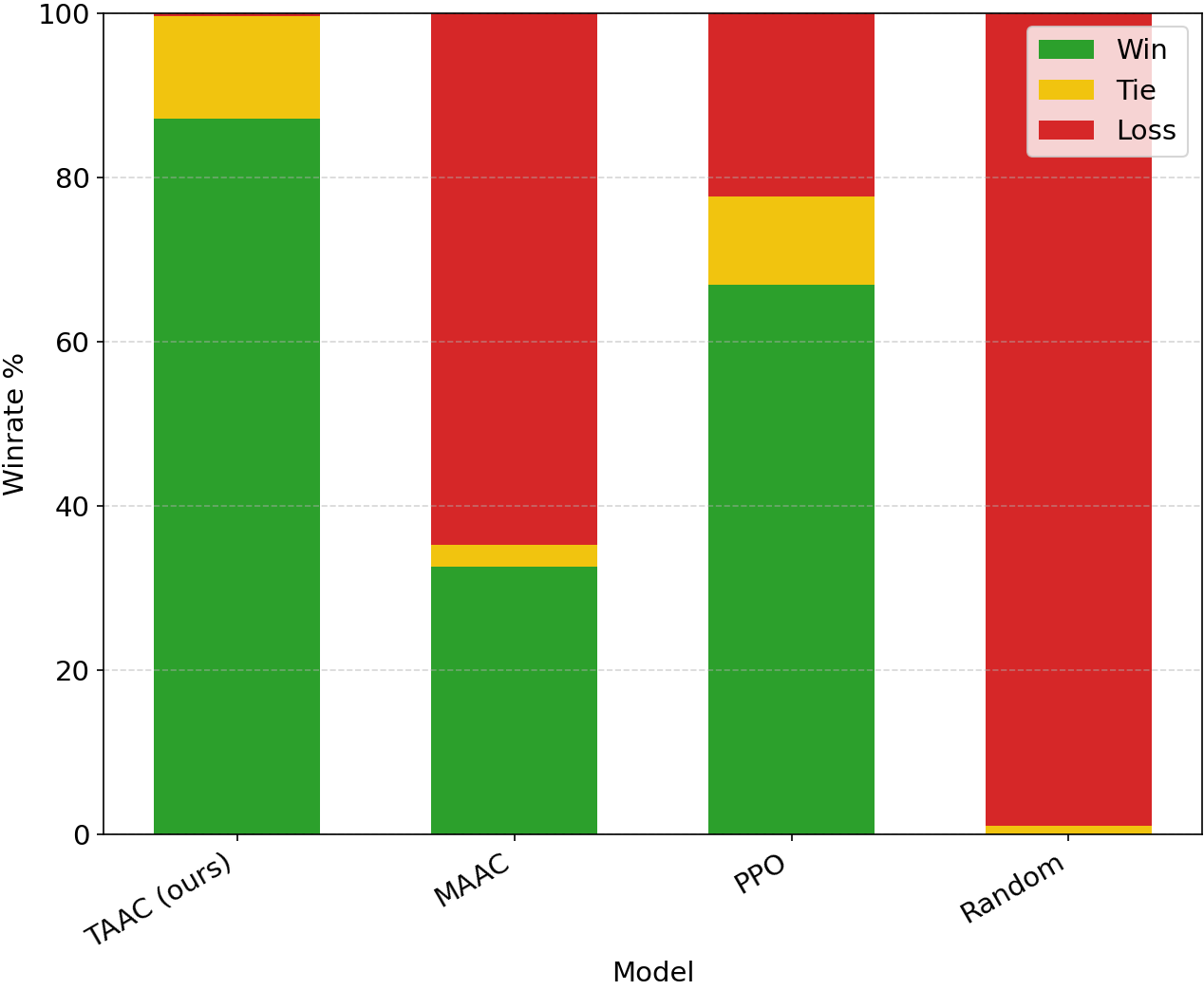}
  \caption{Soccer: This figure showcases the win/tie/loss rate for each algorithm against all others (not including itself).
  It demonstrates TAAC’s better performance when it was paired up head-to-head with the other algorithms.}
  \label{fig:soccer_winrate}
\end{figure}

\begin{figure}[!h]
  \centering
  \includegraphics[width=0.97\linewidth]{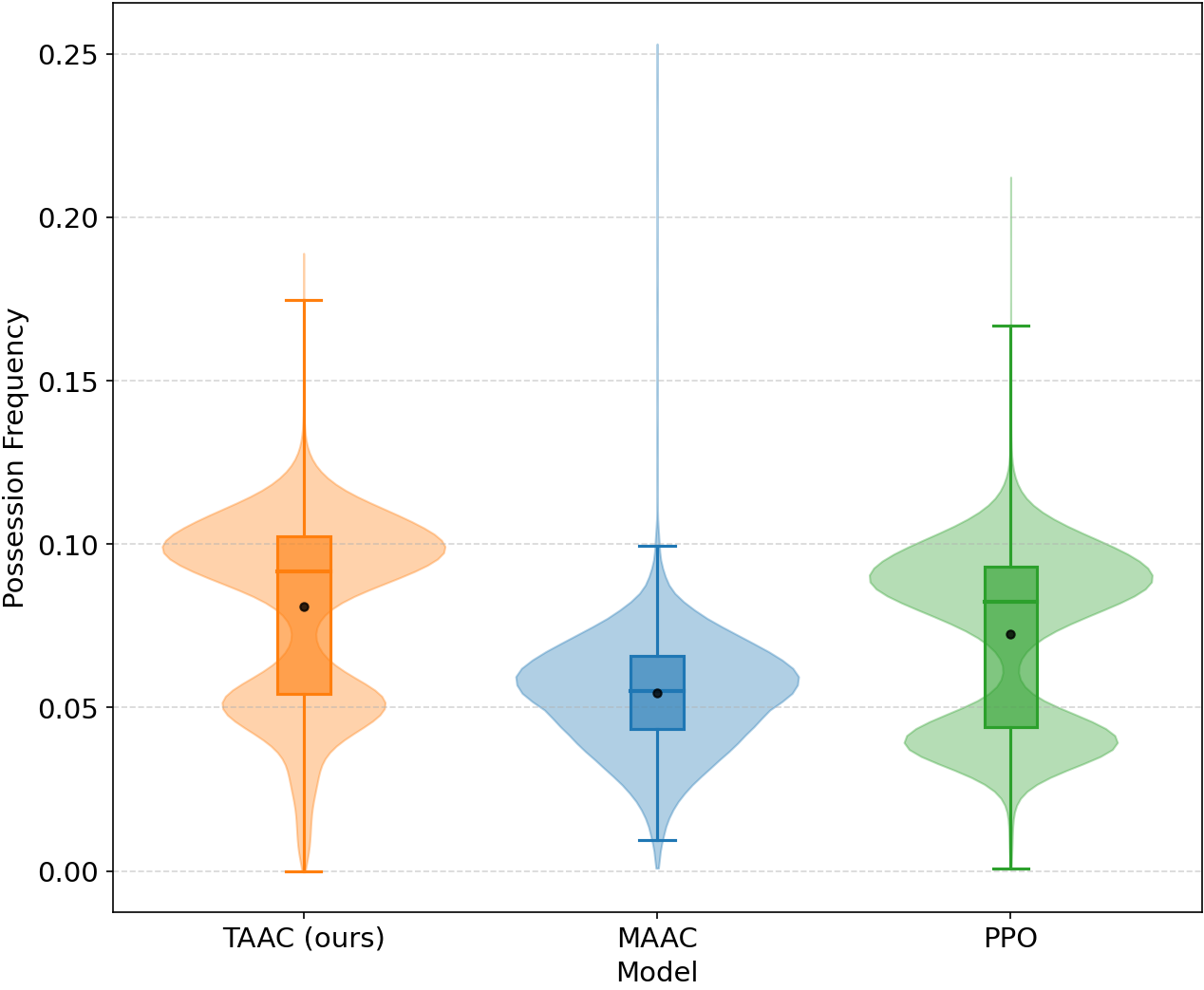}
  \caption{Soccer: Frequency of possession swaps across algorithms over the evaluation period. Higher values may indicate more intra-team passing and collaboration.
  We see how TAAC has a marginally higher frequency of possession swaps, indicating more team passing.
A Kolmogorov-Smirnov Test and a Bootstrap test suggest that all pairs of distributions are statistically different.}
  \label{fig:soccer_possession}
\end{figure}

Figure \ref{fig:soccer_winrate} and Figure \ref{fig:soccer_possession} showcase the performance for each algorithm in the Soccer environment. We hope to see TAAC win games against teams whose players are managed by alternative algorithms and do so with collaboration. In Figure \ref{fig:soccer_winrate} we observe that TAAC's overall win rate is substantially higher than any other approach.  Beyond this, anecdotal testing of TAAC against human players in this environment suggests that TAAC-trained agents have human-level performance.  Figure \ref{fig:soccer_possession} suggests that TAAC-managed players successfully pass significantly more than alternative approaches, indicating more collaboration between team members. We provide more results in this environment in \hyperref[Appendix:C.3]{Appendix C.3}. 

%@@@@@@@@@@@@@@@@@@@@@@@@@@@@@@@@@@@@@@@@@@@@@@@@@
\subsection{Discussion} 
%\noindent Overall, these results suggest that TAAC has better or comparable performance to the other algorithms in cooperative environments and that this advantage becomes more apparent when the task requires more collaboration and coordination. This is clearly demonstrated in the BoxJump environment, where TAAC is superior in a statistical sense above 8 agents (Figure \ref{fig:boxjump_height}) while for 7 agents and below there is no significant difference in performance with MAAC. This is also demonstrated in the soccer environment where TAAC has a higher win rate (Figure \ref{fig:soccer_winrate}) due to its improved ability to coordinate and collaborate in a highly strategic and competitive environment. Across LBF tasks (Figure \ref{fig:lbf_results}), TAAC does not significantly outperform the alternative approaches, but does evidence increasing advantage as the environment becomes more complicated. We attribute this to the relatively low coordination requirement in LBF: success often hinges on simple timing and assignment for food collection rather than sustained, multi-agent interplay. In fact, at certain levels of players and food, a single agent can complete many pickups, limiting the benefit of richer teamwide inter-agent communication. Overall, TAAC is a strong performer in complex MARL environments which require a high level of coordination and collaboration. As shown in the results, outside this domain, TAAC has comparable performance to the other algorithms.

\noindent Overall, these results suggest that TAAC has better or comparable performance to the other algorithms in cooperative environments and that this advantage becomes more apparent when the task requires more collaboration and coordination. 

This is clearly demonstrated in the \textbf{BoxJump} environment, where TAAC is able to achieve a higher height than the other algorithms as seen in \ref{fig:boxjump_height}.  It is also able to do this more quickly with more reward (see Figures \ref{fig:boxjump_steps} and \ref{fig:boxjump_reward} in the appendix). This superiority is related to task complexity in the sense that it grows as the number of agents increases and coordination becomes more difficult; TAAC height is significantly greater than the alternatives above 8 agents (Figure \ref{fig:boxjump_height}) while for 7 agents and below there is no significant difference in performance with MAAC. This showcases how TAAC scales better at coordinating a larger number of agents.

It is also demonstrated in the \textbf{soccer} environment where TAAC has a higher win rate (Figure \ref{fig:soccer_winrate}). It also has a higher Elo rating than the other algorithms (see Figure \ref{fig:soccer_elo_ratings} in the appendix).   We suspect that this is due to its improved ability to coordinate and collaborate in a highly strategic and competitive environment. This ability is on display in Figure \ref{fig:soccer_possession}, which shows that TAAC more frequently passes the ball than the other algorithms. Note that passing requires clear line of sight between players; Figure \ref{fig:soccer_connectivity} in the appendix indicates that TAAC-managed players do not maintain the most passing lanes. Given that TAAC is outscoring and passing more frequently it is likely that it is using its passing lanes more effectively as opposed to MAAC, which has a large amount of passing lanes but very few passes. From our perspective, this suggests that TAAC learns to create and leverage collaborative opportunities dynamically in a selective fashion, only when it is optimal to do so.

We also see the TAAC-managed players are overall closer together on the field than the players managed by alternatives, as seen in Figure \ref{fig:soccer_distance} in the appendix; in an anecdotal sense, from watching recordings of the algorithms, TAAC-players appear to have a very dynamic strategy, where either the whole team is attacking or the whole team is defending, consistent with its low pairwise distance and connectivity. These recordings also showcase how TAAC is passing the ball more frequently and is more offensive than the other algorithms.

Being more spread out could indicate a stronger sense of roles and responsibilities. A goalie, defenders, and attackers are likely to control different parts of the field and therefore be more spread out. This is consistent with our observations of play; the other algorithms were more spread out and had a more role-based approach, often with dynamic defenders and even goalies; while this approach shows some form of collaboration, it did not seem to be the best strategy for coordinating three agents.
Being able to pass the ball around the offense and coordinate an airtight defense seemed to be the best strategy and the one that required the most collaboration.

Finally, across \textbf{LBF} tasks, TAAC's advantage appears and then disappears as the player count varies.  In two-player tasks seen in Figure \ref{fig:lbf_results} top (and Figure \ref{fig:lbf_results_3} in the appendix), we see that TAAC performs as well as the other algorithms. As we increase the players to three, TAAC performs marginally better than the other algorithms (see Figures \ref{fig:lbf_results} bottom and \ref{fig:lbf_results_4} in the appendix). As we increase the number of agents to four and five, TAAC becomes comparable to the other algorithms again (Figures \ref{fig:lbf_results_5} and \ref{fig:lbf_results_6} in the appendix). 

We hypothesize that this is because coordinating two agents to gather food is a relatively low complexity task while coordinating three agents to gather food is more difficult. As team size grows, the LBF task becomes over-provisioned: if a pickup needs $k$ agents, having $n \ge k$ makes accidental coordination likely, limiting the benefit of richer teamwide inter-agent communication and thereby reducing the marginal benefit of TAAC’s collaboration and yielding comparable performance. 

Overall, TAAC is a strong performer in complex MARL environments which require a high level of coordination and collaboration. As shown in the results, outside this domain, TAAC has comparable performance to the other algorithms. We also showcase that TAAC is able to scale better with the number of agents than other algorithms. 

%%%%%%%%%%%%%%%%%%%%%%%%%%%%%%%%%%%%%%%%%%%%%%%%%%%%%%%%%%%%%%%%
\section{Conclusion}  
\noindent In this work, we introduced TAAC, a multi-agent reinforcement learning algorithm that leverages multi-headed attention in both the actor and critic networks to foster robust collaboration. We evaluated TAAC in a diverse range of cooperative environments, to understand the conditionality which allows TAAC to outperform the other algorithms. Our experiments demonstrate that TAAC outperforms other algorithms in MARL environments which require a high level of coordination and collaboration. Future work could focus on:
\begin{enumerate}
    \item comparing against a wider range of algorithms in more environments;
    \item measuring how performance differs with more attention layers;
    \item implementing explainability methods to analyze how the attention mechanism is used by the agents and what information is being shared.
\end{enumerate}

\bibliographystyle{IEEEtran}
\bibliography{references}

%\clearpage
\section{Appendix}

%%%%%%%%%%%%%%%%%%%%%%%%%%%%%%%%%%%%%%%%%%%%%%%%%%%%%%%%%%%%%%%%%%%%%%%%%%%%%%%%%%%%%%%%%%%%%%%%%%%%%
\subsection* {Appendix A : Soccer Environment Description}\label{Appendix:A}
\noindent The soccer simulation is played on a top-down 2D plane; the players and the ball are represented as solid circles on the plane. Players can exert a force on the ball by colliding with it. When a player collides with the ball while in mode ``kicking", the ball gains additional force. Additionally, the ball collides nearly elastically with the walls, preventing it from leaving the play area.  

The observations of the agent are composed of:
\begin{itemize}
    \item The relative position vector to each teammate.
    \item The relative position vector to each agent on the opposing team.
    \item The relative position vector to the ball.
    \item The velocity vector of the ball.
    \item The relative position vector to the agent's own goal.
    \item The relative position vector to the opponent's goal.
    \item Raycast measurements in the directions of North, East, West, and South, indicating the distance to the nearest boundaries of the play area.
\end{itemize}

In our environment, an action is defined by five Boolean inputs: move forward, move right, move down, move left, and kick.
An agent may choose any combination of these inputs, provided they are not contradictory (e.g., moving left and right simultaneously). Overall, the total number of possible actions is 18.

After an action is chosen and the simulation advances by one time step, rewards are assigned to each agent. The reward structure is designed as follows:
\begin{itemize}
    \item \textbf{Exploration Reward:} Each agent receives a small reward for moving toward the ball. This reward is computed as the dot product of the agent's chosen direction $\vec{d}$ and the vector from the agent to the ball $\vec{v}_a$, scaled by a factor $\theta_{\text{exp}}$: 
    \begin{align}
     r_{\text{explore}} = \theta_{\text{exp}} \, (\vec{d} \cdot \vec{v}_a)
    \end{align}
    \item \textbf{Team Reward:} All agents on the same team receive a larger reward for moving the ball towards the opponent's goal. This reward is calculated as the dot product of the ball's velocity vector $\vec{v}_b$ and the vector from the ball to the opponent's goal $\vec{g}$ scaled by a factor $\theta_{\text{ball}}$:
    \begin{align}
     r_{\text{ball}} = \theta_{\text{ball}} \, (\vec{v}_b \cdot \vec{g})
    \end{align}
    \item \textbf{Scoring Reward:} The biggest reward $ r_{\text{goal}}$ is granted when a goal is scored, defined as the ball being fully inside the goal box.
    \item \textbf{Distance Reward:} This reward is allocated based on the average distance to all teammates, to prevent players from clustering together.
    First, the average distance is computed, then it is capped at a maximum threshold, and finally scaled by a constant factor to prevent excessive rewards for large spacing. Specifically, if $d_i$ represents the distance to teammate $i$ and there are $N$ teammates, the average distance is calculated as
    $\bar{d} = \frac{1}{N}\sum_{i=1}^{N} d_i$
    and the reward function is defined as:
    \begin{align}
    f(\bar{d}) = \theta_{\text{dist}} \cdot \min\left(\bar{d},\ \theta_{\text{max}}\right).
    \end{align}
    Here, $\theta_{\text{dist}}$ is the scaling factor and $\theta_{\text{max}}$ is the cap applied to the average distance.
\end{itemize}

It is important to note that all rewards (except for the distance reward) can be negative, acting as penalties for behaviors such as moving away from the ball, moving the ball away from the opponent's goal, or being scored on.
Games are configured to last $T$ time steps, and an episode is defined as a sequence of time steps that terminates either when a goal is scored or when the game concludes. As a result, a single game may consist of multiple episodes.

\vspace{0.8em}
\noindent
\textit{A.1. Evaluation Metrics}
\vspace{0.2em}

\noindent Soccer is a sport where teamwork is not needed to succeed, such as by having a single player who is capable of dribbling and scoring past all of the opposing team. 
Because of this, in this environment we decided to measure both metrics which indicate performance (winning-capability) as well as collaboration.
Performance is evaluated using win rates, goal differentials, and Elo ratings \cite{minka2018trueskill}. 
Each team is assigned an Elo rating, which gets updated based on victory, loss or tie and the Elo rating of the opponent. 
When interpreting Elo ratings, it is important to note that the absolute values are less meaningful than the relative differences between them.
Additionally, goal differential is measured as the difference between the teams' scores.
Collaboration is measured through:
\begin{itemize}
    \item \textbf{Average Pairwise Distance:} A higher average distance suggests that agents are not clustering excessively, which may indicate diverse task allocation.
    \item \textbf{Possession Frequency:} This measures the number of times a ball swaps possession within the team. A high number may indicate that players are passing and, therefore, collaborating.
    \item \textbf{Connectivity:} This metric measures the number of unobstructed graph-like connections between players, calculated as the ratio of observed connections to the maximum possible connections $\frac{N(N-1)}{2}$. 
    To ensure only meaningful connections are counted, only connections within a range of distances are included.
    An optimal connectivity level indicates balanced spacing and collaborative positioning. 
    Intuitively this can be thought of as how many unobstructed passing lanes the team has.
\end{itemize}
Using these metrics, we can comprehensively assess both the performance and collaborative behavior of the teams.

%%%%%%%%%%%%%%%%%%%%%%%%%%%%%%%%%%%%%%%%%%%%%%%%%%%%%%%%%%%%%%%%%%%%%%%%%%%%%%%%%%%%%%%%%%%%%%%%%%%%%
%\clearpage
\subsection*{Appendix B : BoxJump Environment Description}\label{Appendix:B}
\noindent In this environment, agents are 2D 1-meter-square objects in a physics environment. They have the ability to move left and right and jump. 
Their goal is to jump on top of each other to build the tallest tower so as to push the red line as high as possible. 

At each time step, every agent derives an individual observation from the global state. In our environment, the state consists of:
\begin{itemize}
    \item Horizontal/Vertical Position of agent.
    \item Horizontal/Vertical velocity of agent.
    \item Distance to closest box in cardinal directions (maximum distance when no box is in a given direction).
    \item Boolean of whether the box can jump, or if it is mid-air.
    \item An integer representing the number of agents in the environment.
\end{itemize}

Based on its observation, an agent selects an action, causing a force to be exerted on the box.
The agent has three potential actions: move right, move left, and jump, each exerting a force on the box in the given direction.
The episode ends if an agent leaves the playable area or if the highest attainable height of $n$ meters is reached.

The reward structure for an individual agent is:
\begin{align*}
r = \begin{cases}
500\,\text{height}_{\max}^{3}, &
  \shortstack[l]{if contributing to tallest tower \&\\ limit is reached} \\
\text{height}_{\max}^{3}, & \text{if contributing to tallest tower} \\
-100, & \text{if leaving play area} \\
0, & \text{otherwise}
\end{cases}
\end{align*}

where $\text{height}_\text{max} = \max(\text{height}_1, \text{height}_2, \cdots, \text{height}_n)$ 
We chose to cube the max height to further incentivize the agents to build the tallest tower and because the difficulty of adding another box given some height is not linear.
Intuitively, each new additional height requires the coordination of all agents currently on the tower, making it increasingly difficult to add another box.

%%%%%%%%%%%%%%%%%%%%%%%%%%%%%%%%%%%%%%%%%%%%%%%%%%%%%%%%%%%%%%%%%%%%%%%%%%%%%%%%%%%%%%%%%%%%%%%%%%%%%
%\newpage
\subsection* {Appendix C : Extended Results \& Analysis} \label{Appendix:C}

\vspace{0.8em}
\noindent
\textit{C.1. BoxJump Extended Results}
\label{Appendix:C.1}
\vspace{0.2em}

\noindent As well as the results in Figure \ref{fig:boxjump_height}, we also showcase the reward and steps taken to reach a goal height for each algorithm in the BoxJump environment.

\begin{figure}[!h]
  \centering
  \includegraphics[width=0.97\linewidth]{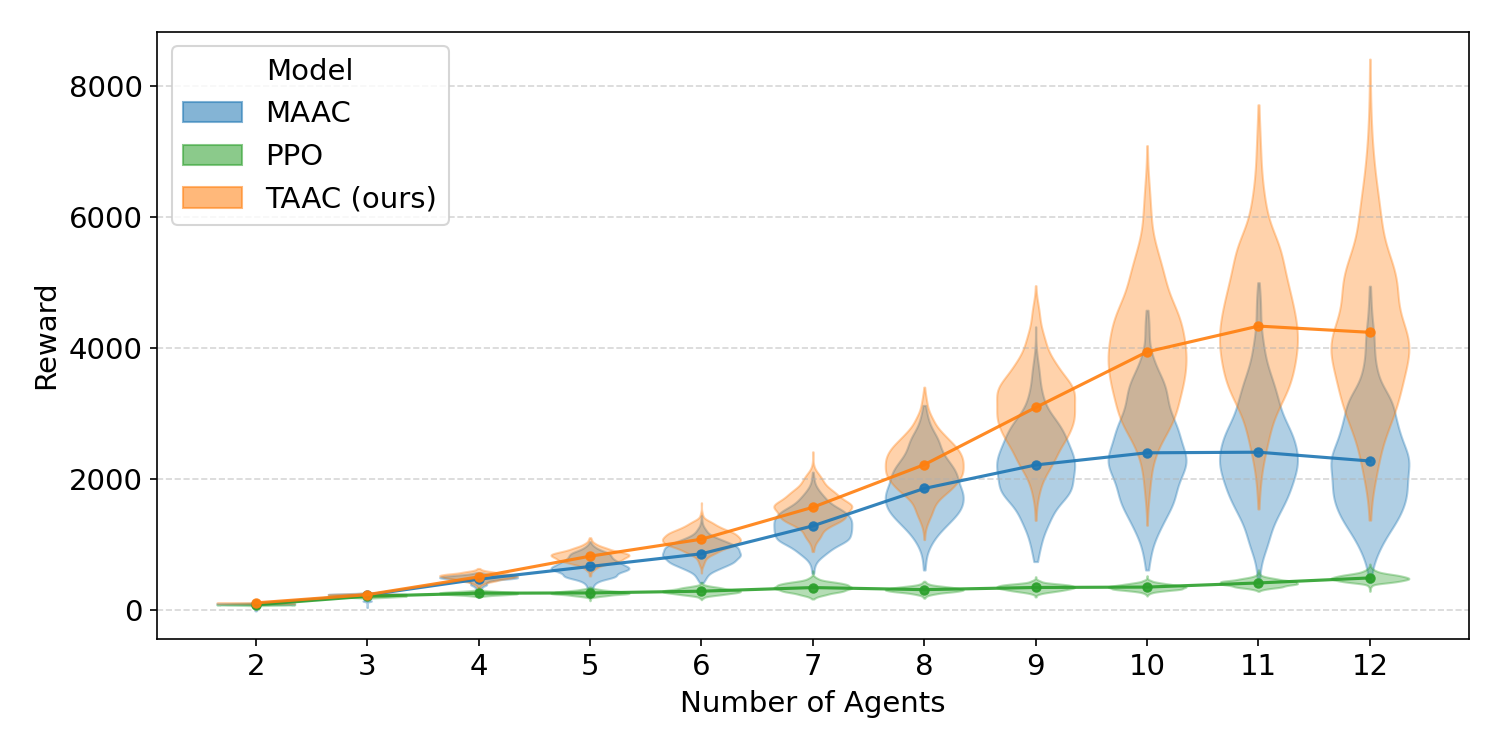}
  \caption{BoxJump: This figure showcases the reward achieved for each agent count and algorithm during inference.
   We can see how TAAC (Ours) achieves a better reward as the number of agents increases.
   Reward is measured as the cube of the maximum height achieved, so to optimize reward agents need to build the tallest tower as fast as possible and stay in that position for as long as possible.
   A Kolmogorov-Smirnov Test and a Bootstrap test suggest that all pairs of distributions are statistically different.}
  \label{fig:boxjump_reward}
\end{figure}

\begin{figure}[!h]
  \centering
  \includegraphics[width=0.97\linewidth]{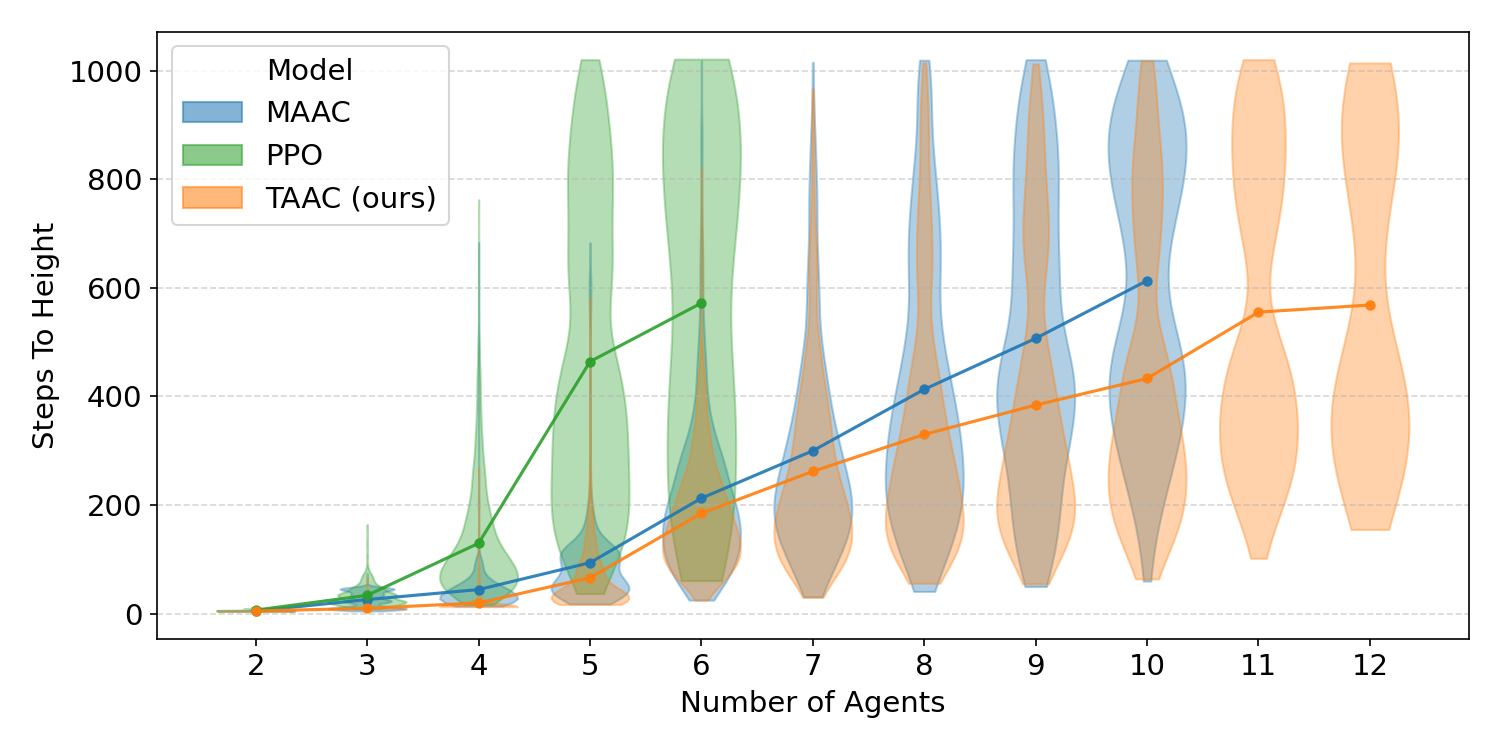}
  \caption{BoxJump: This figure showcases the number of steps taken to reach a goal height of $n$ meters where $n$ is the number of agents.
  We can see how TAAC (Ours) takes fewer steps to achieve its perfect goal and additionally is capable of reaching the goal height at all for higher numbers of agents.}
  \label{fig:boxjump_steps}
\end{figure}

\newpage
\vspace{0.8em}
\noindent
\textit{C.2. LBF Extended Results}
\label{Appendix:C.2}
\vspace{0.2em}

\begin{figure}[!h]
  \centering
  \includegraphics[width=0.97\linewidth]{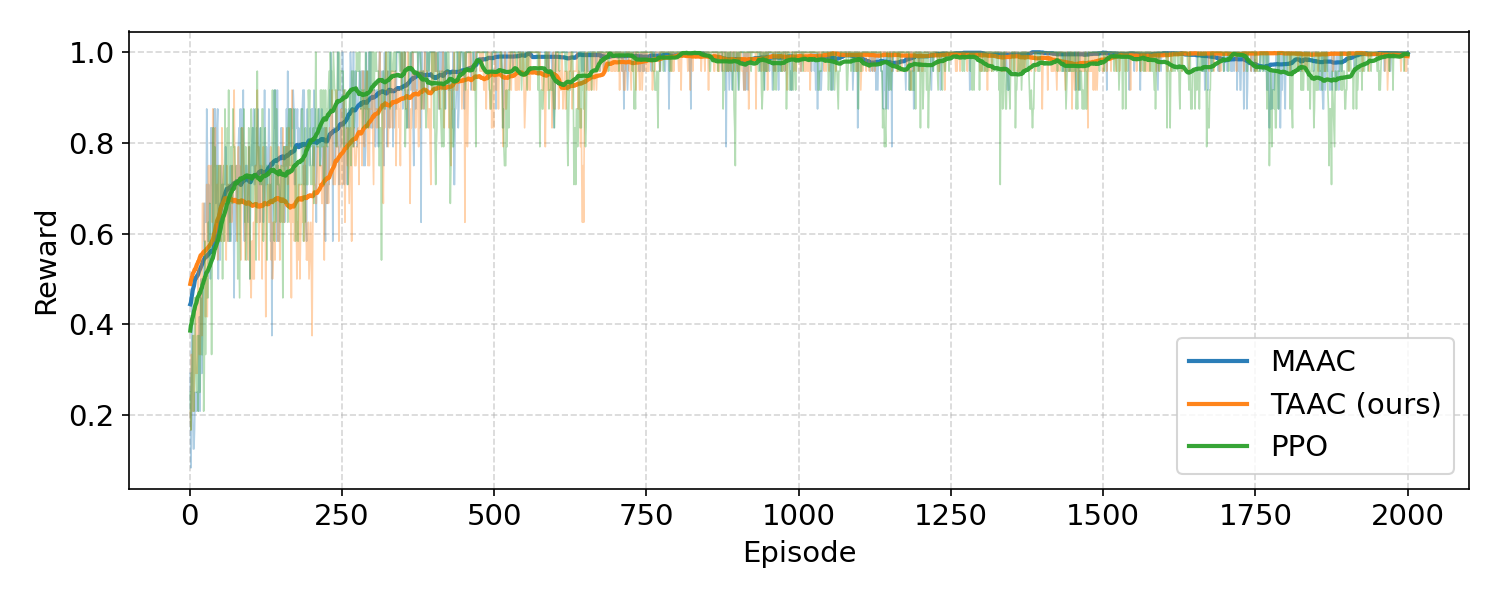}
  \caption{LBF - Task: 2p-2f-5x5-c.
  2 players gather 2 food in a 5 by 5 grid world; each gathering requires all players.
  Reward is normalized by food levels.}
  \label{fig:lbf_results_3}
\end{figure}

\begin{figure}[!h]
  \centering
  \includegraphics[width=0.97\linewidth]{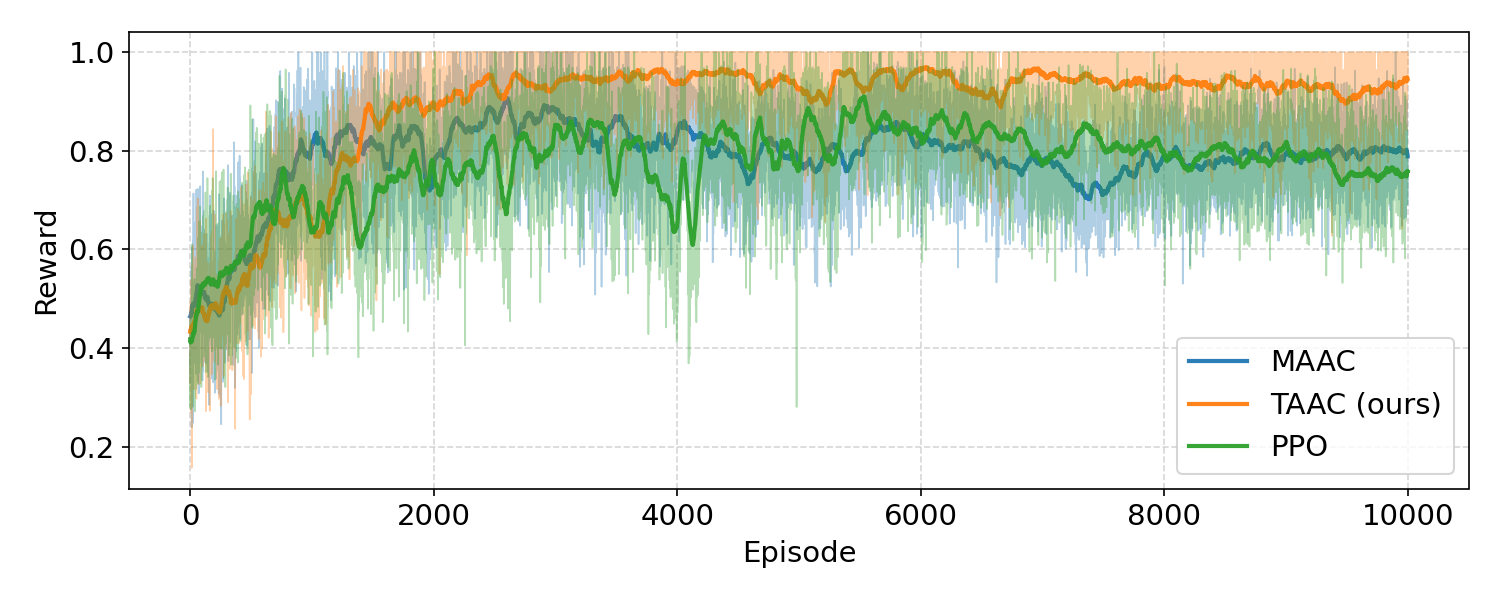}
  \caption{LBF - Task: 3p-3f-6x6
  3 players gather 3 food in a 6 by 6 grid world.
  Reward is normalized by food levels.}
  
  \label{fig:lbf_results_4}
\end{figure}

\begin{figure}[!h]
  \centering
  \includegraphics[width=0.97\linewidth]{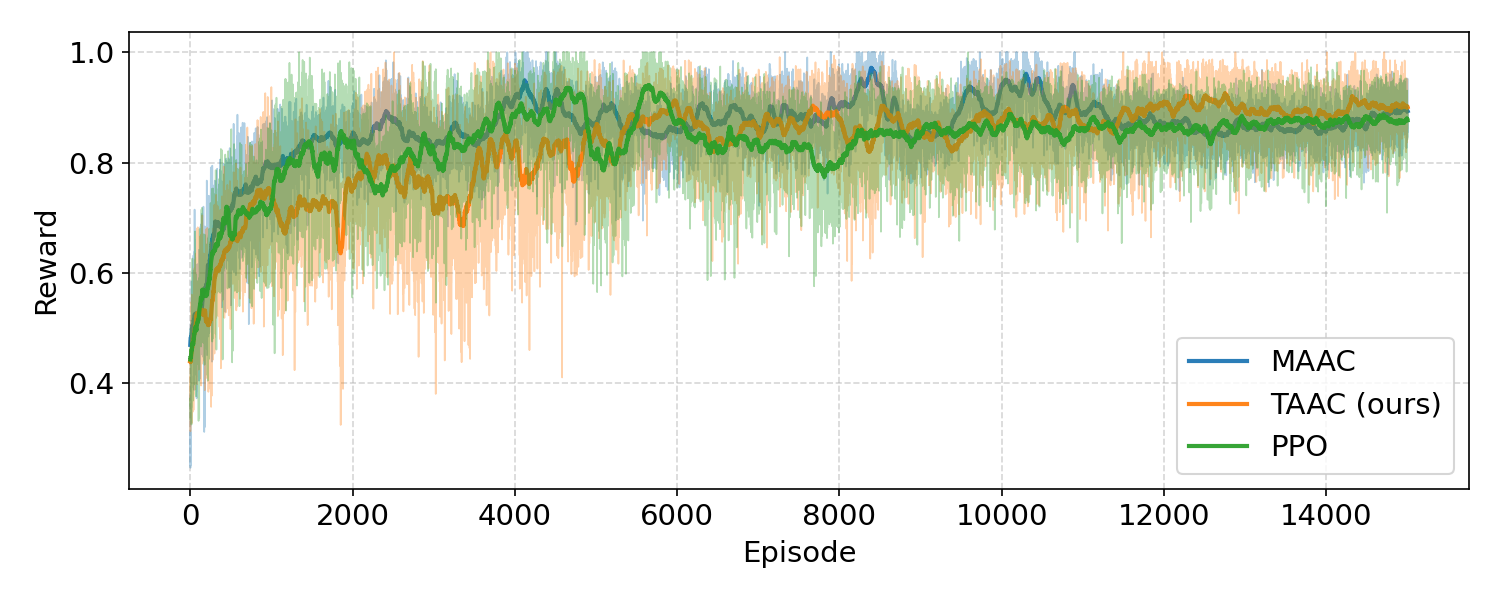}
  \caption{LBF - Task: 4p-8f-8x8
  4 players gather 8 food in an 8 by 8 grid world.
  Reward is normalized by food levels.}
  \label{fig:lbf_results_5}
\end{figure}

\begin{figure}[!h]
  \centering
  \includegraphics[width=0.97\linewidth]{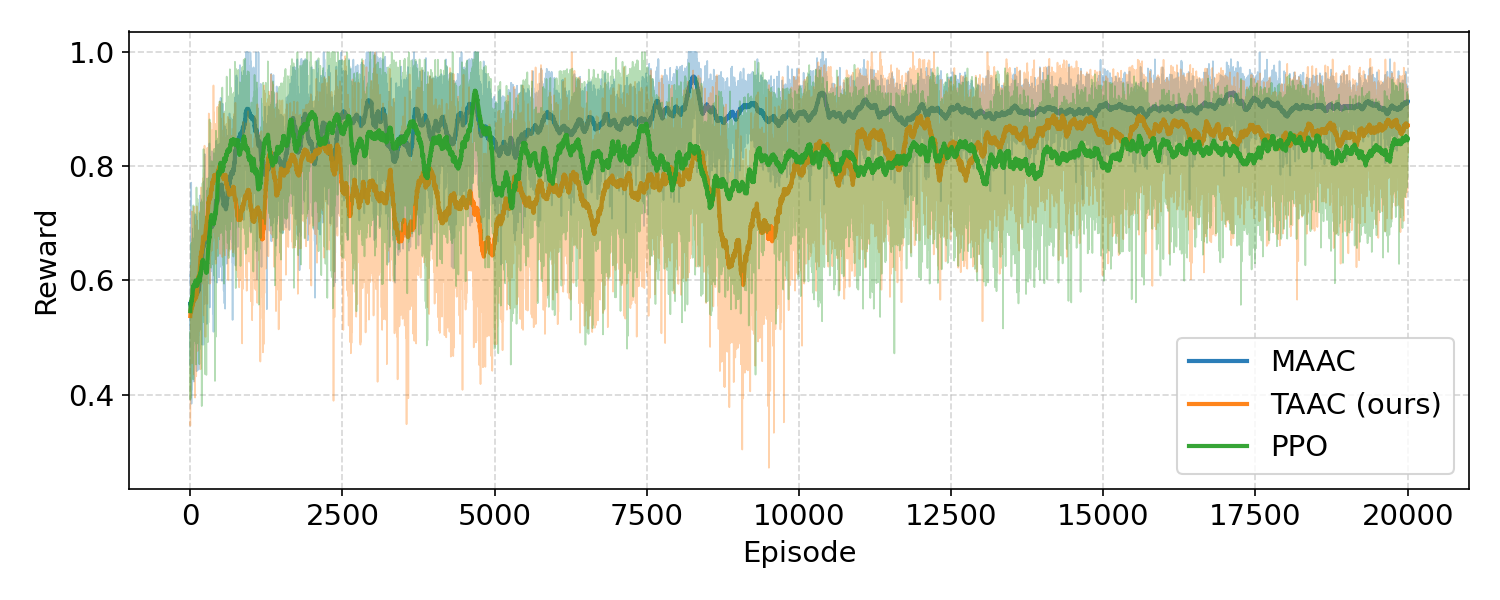}
  \caption{LBF - Task: 5p-10f-9x9
  5 players gather 10 food in a 9 by 9 grid world.
  Reward is normalized by food levels.}
  \label{fig:lbf_results_6}
\end{figure}

For LBF, we performed a range of different tasks and gathered the reward through the training process.
In Table \ref{tab:lbf_results}, we showcase the reward for each algorithm for each task.
We also plot the reward for each algorithm for each task in Figures \ref{fig:lbf_results}, \ref{fig:lbf_results_3}, \ref{fig:lbf_results_4}, \ref{fig:lbf_results_5}, \ref{fig:lbf_results_6}.

\begin{table}[!h]
  \caption{LBF tasks: mean $\pm$ 95\% confidence interval reward over the last 500 episodes of training} 
  \label{tab:lbf_results}
  \centering
  \begin{tabular}{lccc}
    \toprule
    \textbf{Task} & \textbf{MAAC} & \textbf{TAAC (Ours)} & \textbf{PPO} \\
    \midrule
    2p-2f-5x5-c & $0.996 \pm 0.026$ & $0.998 \pm 0.018$ & $0.985 \pm 0.050$ \\
    2p-4f-5x5-c & $0.998 \pm 0.022$ & $0.995 \pm 0.032$ & $0.994 \pm 0.033$ \\
    3p-3f-6x6 & $0.798 \pm 0.136$ & $\bm{0.925 \pm 0.124}$ & $0.754 \pm 0.138$ \\
    3p-6f-6x6 & $0.899 \pm 0.113$ & $\bm{0.955 \pm 0.103}$ & $0.764 \pm 0.132$ \\
    4p-8f-8x8 & $0.888 \pm 0.067$ & $\bm{0.902 \pm 0.078}$ & $0.877 \pm 0.074$ \\
    5p-10f-10x10 & $\bm{0.903 \pm 0.052}$ & $0.871 \pm 0.112$ & $0.834 \pm 0.114$ \\
    \bottomrule
  \end{tabular}
\end{table}

\newpage
\vspace{-3.8em}
\noindent
\textit{C.3. Soccer Extended Results}
\label{Appendix:C.3}
\vspace{0.2em}

\begin{figure}[!h]
  \centering
  \includegraphics[width=0.97\linewidth]{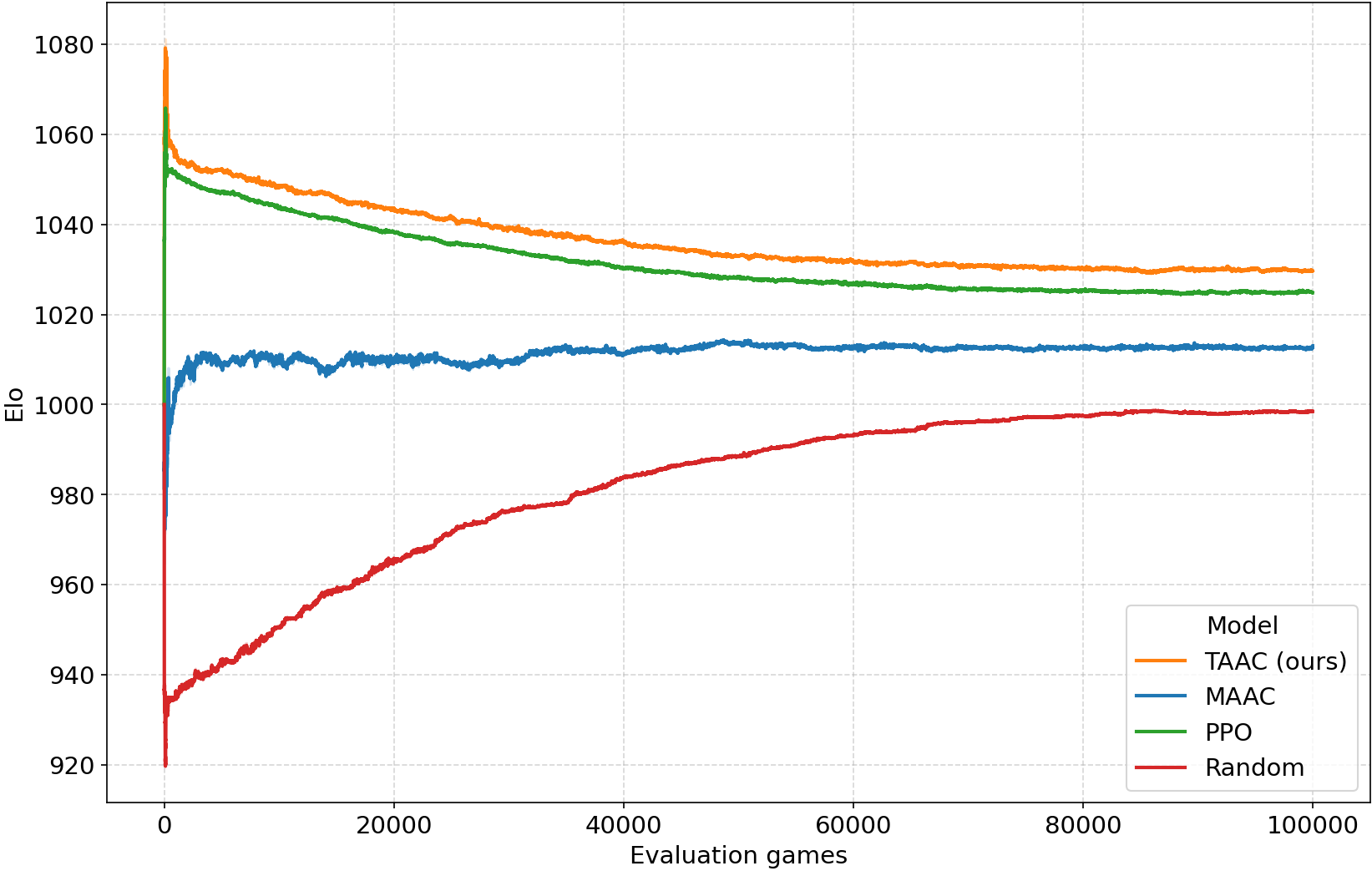}
  \caption{Soccer: Elo ratings for each algorithm over the evaluation period. Elo ratings are a measure of relative skill between two algorithms. TAAC (Ours) achieves a higher elo rating showcasing its superior performance.}

  \label{fig:soccer_elo_ratings}
\end{figure}

\begin{figure}[!h]
  \centering
  \includegraphics[width=0.97\linewidth]{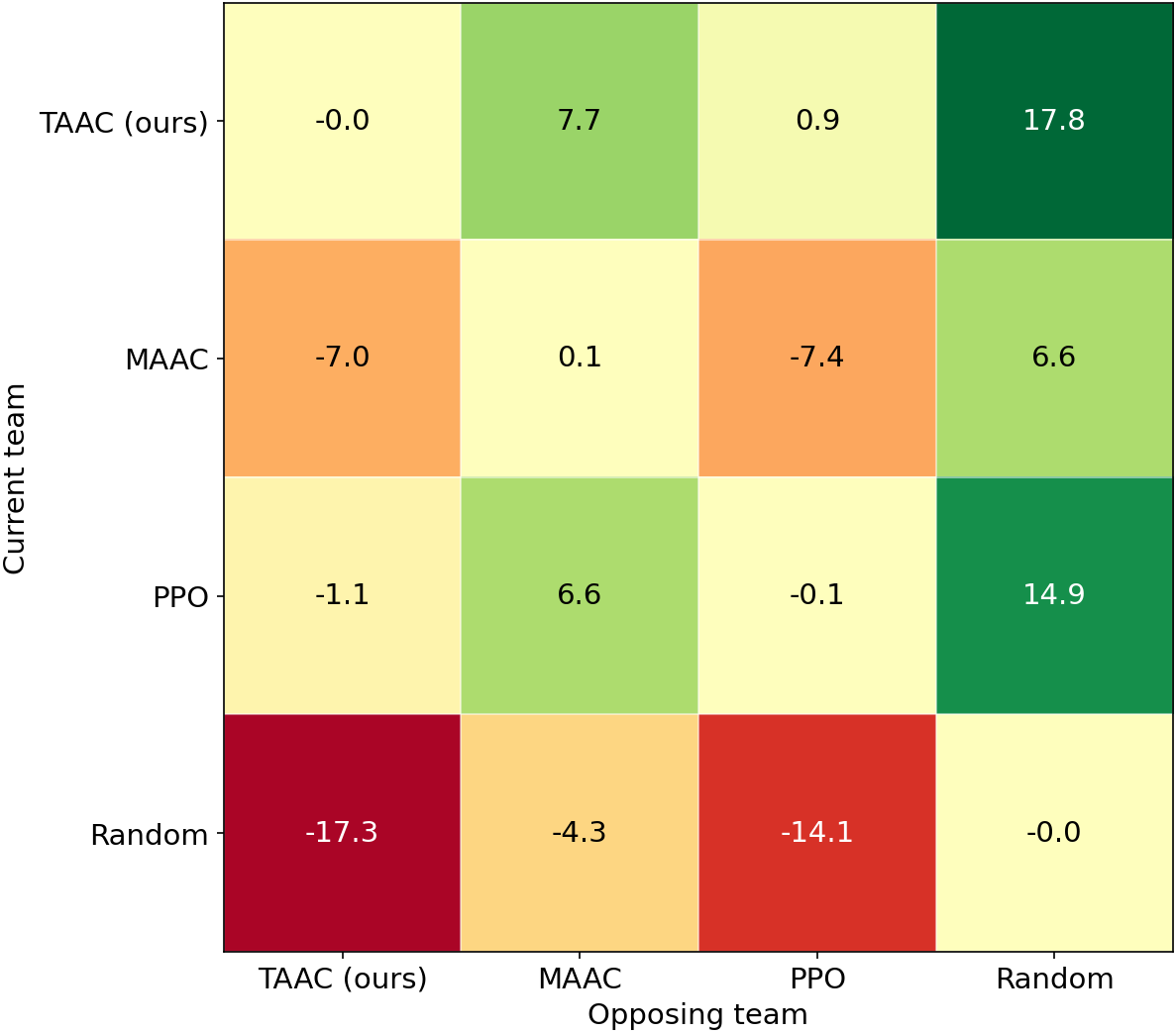}
  \caption{Soccer: Goal differential for each algorithm. 
  Each goal differential is the average difference in goals scored between the team on the column and the team on the row.
  A positive value indicates that the team on the column is outscoring the team on the row, while a negative value indicates the opposite.
  TAAC (Ours) achieves a positive goal differential against all other algorithms, showcasing its superiority across all other algorithms.}
  \label{fig:soccer_goal_diff}
\end{figure}

\begin{figure}[!h]
  \centering
  \includegraphics[width=0.97\linewidth]{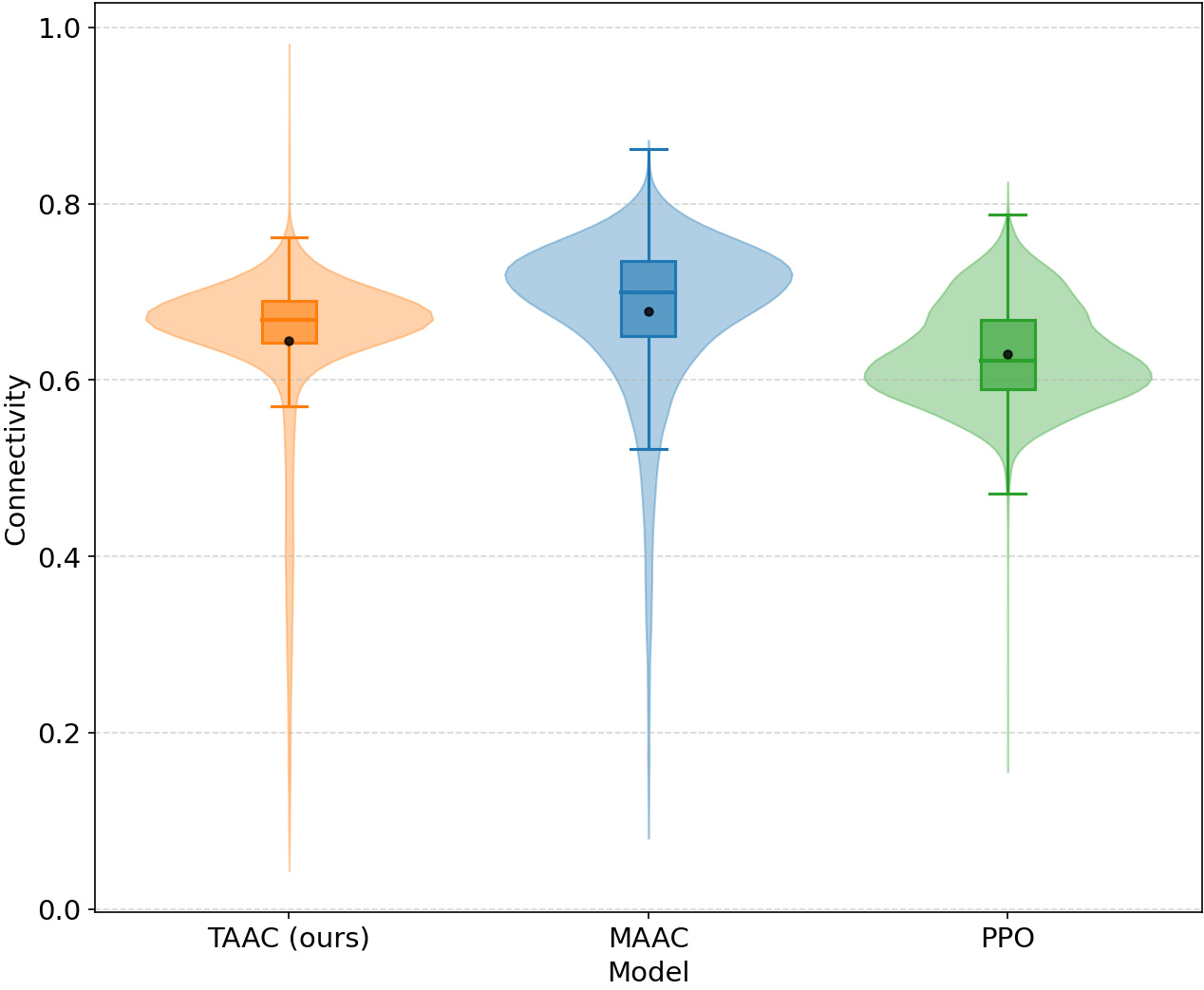}
  \caption{Soccer: Connectivity for each algorithm.
  As described in \hyperref[Appendix:A]{Appendix A}, connectivity is measured as the ratio of observed connections to the maximum possible connections $\frac{N(N-1)}{2}$.
  It intuitively measures the number of open passing lanes.
  This graph showcases that MAAC has the most, followed by TAAC (ours) and PPO.
  Kolmogorov-Smirnov Test and a Bootstrap test suggest that all pairs of distributions are statistically different.}
  \label{fig:soccer_connectivity}
\end{figure}

\begin{figure}[!h]
  \centering
  \includegraphics[width=0.97\linewidth]{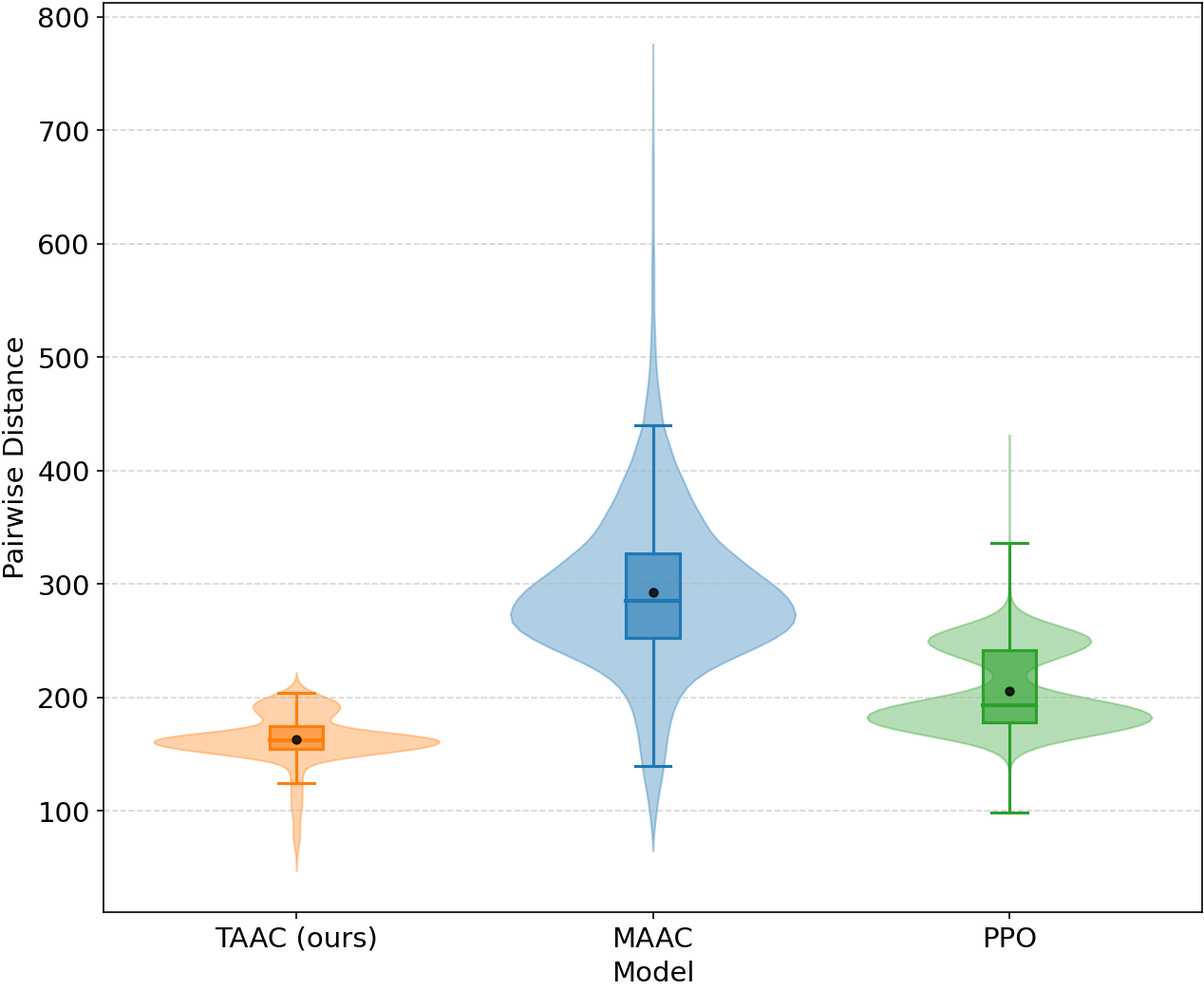}
  \caption{Soccer: Average pairwise distance for each algorithm.
  As described in \hyperref[Appendix:A]{Appendix A}, the average pairwise distance is measured as the average distance between all pairs of agents on the same team.
  TAAC (Ours) achieves a lower average pairwise distance than the other algorithms, showcasing that it is less spread out than the other algorithms.
  Kolmogorov-Smirnov Test and a Bootstrap test suggest that all pairs of distributions are statistically different.}
  \label{fig:soccer_distance}
\end{figure}

\noindent The following are additional results for the soccer environment as described in \hyperref[Appendix:A]{Appendix A}.
We gathered these metrics through the evaluation process, and they can be seen in figures \ref{fig:soccer_elo_ratings}, \ref{fig:soccer_goal_diff}, \ref{fig:soccer_connectivity}, \ref{fig:soccer_distance}.
You can see the win rate and possession frequency in Figures \ref{fig:soccer_winrate} and \ref{fig:soccer_possession} in the main results section.

\end{document}